\documentclass{article}

\usepackage{microtype}
\usepackage{graphicx}
\usepackage{subcaption}
\usepackage{booktabs} 

\usepackage{hyperref}

\usepackage[preprint]{icml2026}

\usepackage[utf8]{inputenc} %
\usepackage[T1]{fontenc}    %
\usepackage{url}            %
\usepackage{booktabs}       %
\usepackage{amsfonts}       %
\usepackage{nicefrac}       %
\usepackage{microtype}      %
\usepackage{xcolor}         %

\usepackage{bbding}
\usepackage{multirow}
\usepackage{algorithm}
\usepackage{graphicx, amsmath, amssymb, caption, subcaption, overpic, textpos}
\usepackage{arydshln} %
\usepackage{listings}
\usepackage{setspace}

\usepackage{xspace}

\newcommand{\tabref}[1]{Tab.~\ref{#1}}

\makeatletter
\DeclareRobustCommand\onedot{\futurelet\@let@token\@onedot}
\def\@onedot{\ifx\@let@token.\else.\fi}
\makeatother

\def\eg{\textit{e.g}\onedot} 
\def\ie{\textit{i.e}\onedot} 
 
 \def\vs{\textit{vs}\onedot}

\newcommand{\figref}[1]{Fig.~\ref{#1}}
\newcommand{\secref}[1]{Sec.~\ref{#1}}

\newlength\savewidth\newcommand\shline{\noalign{\global\savewidth\arrayrulewidth
  \global\arrayrulewidth 1pt}\hline\noalign{\global\arrayrulewidth\savewidth}}
\newcommand{\tablestyle}[2]{\setlength{\tabcolsep}{#1}\renewcommand{\arraystretch}{#2}\centering\footnotesize}

\usepackage{pifont}
\definecolor{mygreen}{RGB}{0,200,0}
\usepackage{colortbl}
\definecolor{finalcolor}{gray}{.9}

\definecolor{baselinecolor}{gray}{.9}
\newcommand{\baseline}[1]{\cellcolor{baselinecolor}{#1}}
\usepackage{subcaption}

\newcommand{\modelname}{BAR\xspace}

\usepackage{amsmath}
\usepackage{amssymb}
\usepackage{mathtools}
\usepackage{amsthm}
\usepackage{adjustbox}   

\usepackage[capitalize,noabbrev]{cleveref}

\theoremstyle{plain}

\theoremstyle{definition}

\theoremstyle{remark}

\usepackage[textsize=tiny]{todonotes}

\icmltitlerunning{Autoregressive Image Generation with Masked Bit Modeling}

\begin{document}

\twocolumn[
  \icmltitle{Autoregressive Image Generation with Masked Bit Modeling}



  \icmlsetsymbol{equal}{*}

  \begin{icmlauthorlist}
    \icmlauthor{Qihang Yu}{comp}
    \icmlauthor{Qihao Liu}{comp}
    \icmlauthor{Ju He}{comp}
    \icmlauthor{Xinyang Zhang}{comp}
    \icmlauthor{Yang Liu}{comp}
    \icmlauthor{Liang-Chieh Chen}{comp2,equal}
    \icmlauthor{Xi Chen}{comp,equal}
    \vspace{4ex}
  \end{icmlauthorlist}

 \icmlaffiliation{comp}{Amazon FAR (Frontier AI \& Robotics)}
 \icmlaffiliation{comp2}{Work done while at FAR. *: Equal advising}
  \icmlcorrespondingauthor{Qihang Yu}{yuqiha@amazon.com}


]



\printAffiliationsAndNotice{}  

\begin{abstract}
This paper challenges the dominance of continuous pipelines in visual generation.
We systematically investigate the performance gap between discrete and continuous methods. Contrary to the belief that discrete tokenizers are intrinsically inferior, we demonstrate that the disparity arises primarily from the total number of bits allocated in the latent space (\ie, the compression ratio). We show that scaling up the codebook size effectively bridges this gap, allowing discrete tokenizers to match or surpass their continuous counterparts.
However, existing discrete generation methods struggle to capitalize on this insight, suffering from performance degradation or prohibitive training costs with scaled codebook. To address this, we propose \textit{masked \textbf{B}it \textbf{A}uto\textbf{R}egressive modeling (\textbf{BAR})}, a scalable framework that supports arbitrary codebook sizes. By equipping an autoregressive transformer with a \textit{masked bit modeling} head, BAR predicts discrete tokens through progressively generating their constituent bits.
BAR achieves a new state-of-the-art gFID of \textbf{0.99} on ImageNet-256, outperforming leading methods across both continuous and discrete paradigms, while significantly reducing sampling costs and converging faster than prior continuous approaches. Project page is available at \url{https://bar-gen.github.io/}

\end{abstract}

\begin{figure}[t!]
    \centering
    \includegraphics[width=0.9\linewidth]{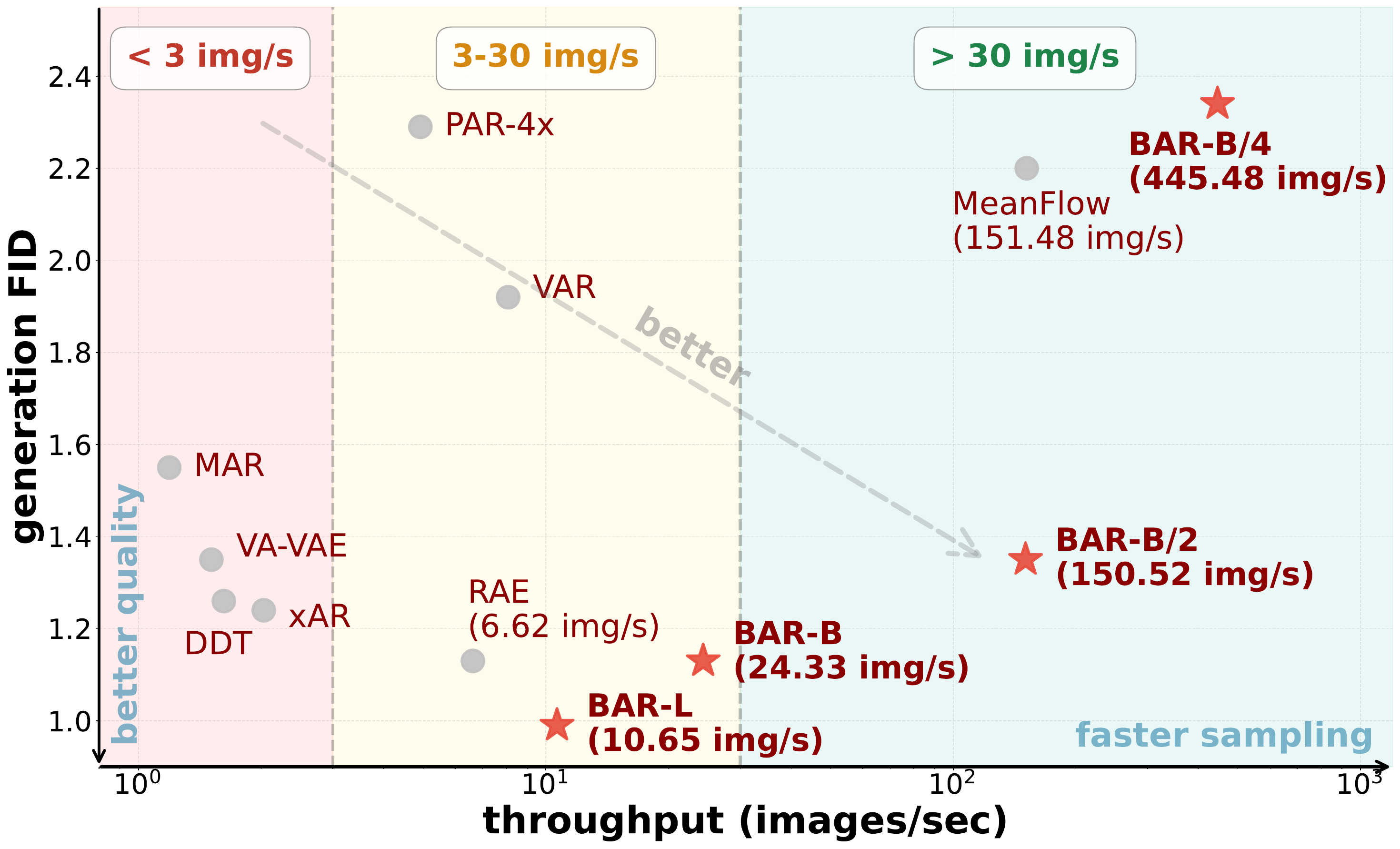}
    \caption{
    \textbf{The proposed \modelname achieves a superior quality-cost trade-off (generation FID \vs \ throughput) on ImageNet-256.} 
    }
    \label{fig:fid}
\end{figure}

\vspace{-3ex}
\section{Introduction}
\label{sec:intro}

Visual generative models have driven remarkable progress across a wide range of computer vision tasks~\cite{wang2023seggpt,team2024chameleon,cui2025emu3,deng2025emerging,wiedemer2025video,agarwal2025cosmos}.
A central component of these systems is visual tokenization, which compresses high-dimensional pixel inputs into compact latent representations. Operating on these latent tokens, a generative model learns the underlying image distribution to synthesize high-fidelity visual content.

Depending on quantization and regularization strategies, visual tokenization and generation pipelines can be broadly categorized into \textit{discrete} and \textit{continuous} approaches.
Each paradigm offers distinct advantages: discrete tokenizers align naturally with language modeling, making them suitable for native multimodal large language models~\cite{team2024chameleon,cui2025emu3}, whereas continuous tokenizers excel at modeling raw visual signals and preserving fine-grained details. Despite progress in both directions, continuous tokenizers, typically with diffusion models, remain dominant in visual generation~\cite{rombach2022high,peebles2023scalable,li2024autoregressive,zheng2025diffusion}. This dominance is largely attributed to their higher information capacity, which enables superior reconstruction fidelity and a higher ceiling for generation~\cite{li2024autoregressive,wang2025bridging}.

In this work, we investigate the performance gap between discrete and continuous pipelines. Our key observation is that this gap is not intrinsically caused by the nature of the representations, but is instead largely associated with differences in the compression rates used in practice. To make this comparison explicit, we unify both paradigms under a common metric: the number of bits used to represent the latent space. From this unified perspective, we find that the commonly observed inferior performance of discrete tokenizers is largely attributable to their substantially higher compression ratios, which lead to severe information loss. Empirically, we show that allocating more bits per token (equivalent to scaling up the codebook size) allows discrete tokenizers to match, and in some cases surpass, their continuous counterparts in reconstruction quality.

While increasing the codebook size narrows the reconstruction performance gap, it poses a significant challenge for generative modeling. Discrete generators are typically trained with cross-entropy objectives, and large vocabularies substantially increase both computational and statistical complexity. In particular, scaling the codebook size makes training prohibitively memory-intensive~\cite{han2025infinity} and increasingly difficult to optimize~\cite{yu2023language}.
To address this, we propose replacing the standard linear prediction head with a lightweight \textit{\textbf{bit generation mechanism}}. Instead of classifying over a massive vocabulary, our method predicts discrete tokens by progressively generating their constituent bits.
This design effectively accommodates unbounded vocabulary sizes and consistently improves generation performance, particularly as the codebook size scales.

In summary, discrete tokenizers can serve as competitive visual compressors relative to their continuous counterparts, and that discrete generators can outperform diffusion models in generation fidelity while achieving faster convergence and higher sampling throughput. Building on these findings, we propose \textit{masked \textbf{B}it \textbf{A}utoreg\textbf{R}essive modeling (\textbf{BAR})}, a strong discrete visual generation framework that challenges the prevailing dominance of continuous pipelines.
\modelname establishes a new state of the art: with only $415$M parameters, it achieves a gFID of $1.13$ on ImageNet-256~\cite{deng2009imagenet}, surpassing prior discrete models while being $3.68\times$ faster than leading continuous approaches~\cite{zheng2025diffusion}. Additionally, our efficient variant matches the performance of one-step model MeanFlow~\cite{geng2025mean} with a $2.94\times$ speedup. Finally, our best-performing variant attains a gFID of $0.99$, setting a new benchmark across both discrete and continuous paradigms.

\begin{figure}[t]
    \centering
    \includegraphics[width=0.9\linewidth]{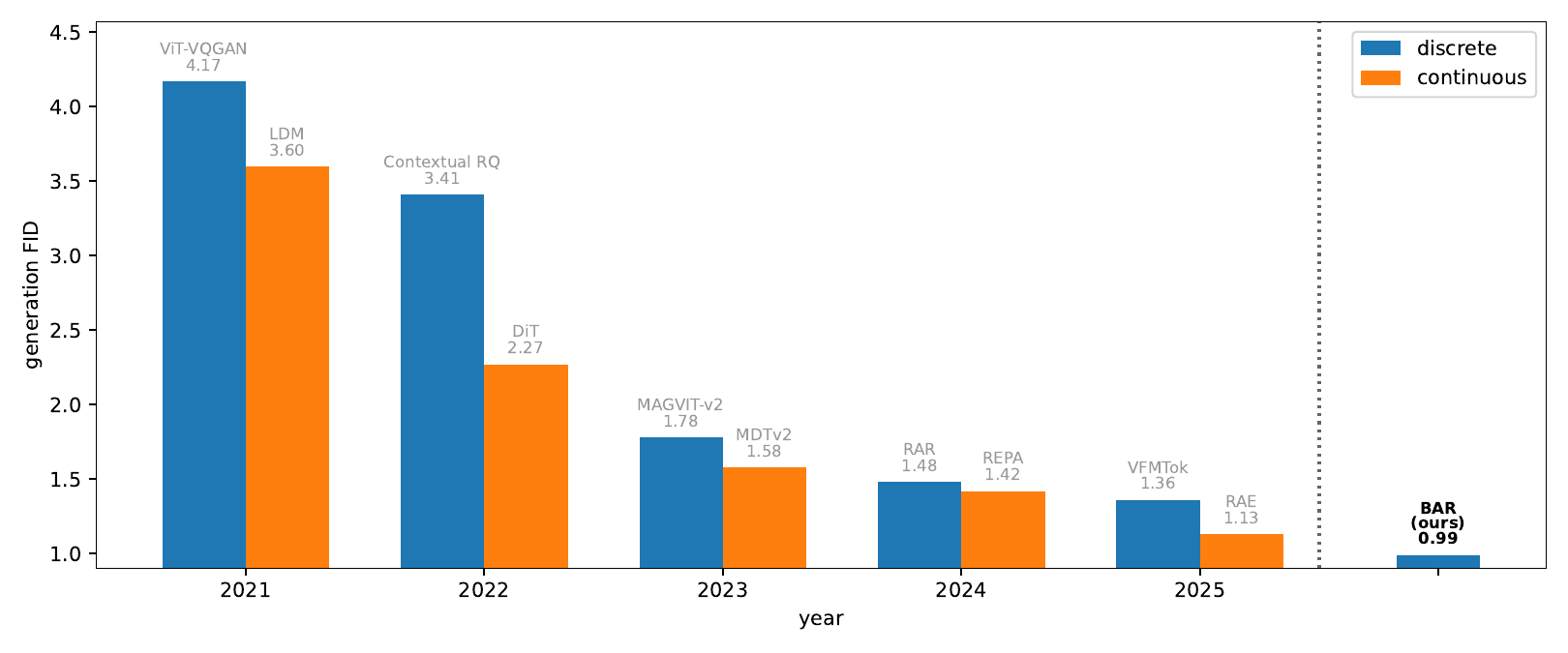}
    \caption{
    \textbf{Best discrete and continuous generator comparison.}
    }
    \vspace{-3ex}
    \label{fig:discrete_continuous_compare}
\end{figure}

\vspace{-1.5ex}
\section{Related Work}

\vspace{-0.5ex}
\noindent \textbf{Continuous Visual Tokenization and Generation.} 
Continuous visual tokenization and generation pipelines typically consist of two main components: a variational autoencoder (VAE)~\cite{kingma2013auto} and a diffusion model~\cite{sohl2015deep,song2019generative,ho2020denoising}. VAEs are autoencoders trained with specific regularization on the latent space (\eg, KL regularization~\cite{rombach2022high}). They usually downsample visual inputs along spatial dimensions while expanding channel dimensions, thereby providing a more compact and structured representation space that is well suited for diffusion-based generation. While a large body of work has focused on diffusion model architectures~\cite{peebles2023scalable,bao2023all,gao2023masked,liu2024alleviating,wang2025ddt}, denoising trajectories~\cite{lipman2022flow,ma2024sit,liu2025flowing}, and prediction objectives~\cite{li2024autoregressive,ren2024flowar,ren2025beyond,he2025flowtok}, SD-VAE~\cite{rombach2022high} has remained the de facto standard VAE backbone in most studies. More recently, increasing attention has been paid to enriching the semantic content of VAE latent spaces, either by incorporating off-the-shelf models~\cite{yao2025reconstruction} or by using frozen encoders as tokenizers~\cite{zheng2025diffusion}. There are also works~\cite{hoogeboom2024simpler,li2025fractal,li2025back} that explore tokenizer-free diffusion models operating in pixel space.

\vspace{-0.5ex}
\noindent \textbf{Discrete Visual Tokenization and Generation.}
Building on the foundation of VQGAN~\cite{esser2021taming}, a substantial body of work has focused on quantizer, the core component of discrete pipelines.
One stream of research aims to enhance the utilization and training dynamics of vanilla vector quantization with learnable codebooks~\cite{yu2021vector,zheng2023online,zhu2024scaling}. Conversely, other approaches abandon learnable codebooks entirely in favor of ``lookup-free'' quantizers~\cite{mentzer2023finite,yu2023language,zhao2024image}.
Notably, while these approaches tokenize images into ``bit tokens,'' they primarily emphasize the benefits of lookup-free quantization, and do not exploit this bit-level structure to redefine the generation targets.

Among these studies, the most closely related works are MaskBit~\cite{weber2024maskbit} and Infinity~\cite{han2025infinity}. MaskBit~\cite{weber2024maskbit} adopts LFQ~\cite{yu2023language} as the tokenizer and directly feeds bit tokens into the generator. However, it still predicts codebook indices rather than bits during generation, which limits scalability with respect to codebook size, similar to standard discrete generative models. Infinity~\cite{han2025infinity} scales to extremely large codebook sizes ($2^{64}$) using BSQ~\cite{zhao2024image} and directly generates images from bits. Nevertheless, it relies heavily on the VAR generator~\cite{tian2024visual} and an external bit-corrector as a post-processing module. In contrast, the proposed BAR framework is compatible with arbitrary autoregressive formulations and generates bit tokens correctly in a fully self-contained manner, enabled by the proposed masked bit modeling head.

\section{Method}

\vspace{-0.5ex}
\subsection{Background}
\vspace{-0.5ex}
We begin by introducing the visual tokenization process. A visual tokenizer, whether discrete or continuous, can be viewed as an autoencoder~\cite{hinton2006reducing} equipped with an information bottleneck~\cite{kingma2013auto,esser2021taming,mentzer2023finite}. Structurally, it consists of three key components: an encoder \(Encoder\), a bottleneck module \(Bottleneck\), and a decoder \(Decoder\). 
The nature of the bottleneck distinguishes the two paradigms: a \textit{discrete} bottleneck maps latent features to entries in a finite codebook, whereas a \textit{continuous} bottleneck typically employs dimensionality reduction coupled with regularization, such as the KL-divergence penalty.

Given an input image \(\mathbf{I} \in \mathbb{R}^{H \times W \times 3}\), where \(H\) and \(W\) denote the image height and width, respectively, the encoder first maps the image to a dense feature map $\mathbf{L}$:
\begin{equation}
    \mathbf{L} = Encoder(\mathbf{I}),
\end{equation}
where \(\mathbf{L}\) is the encoded feature with spatial shape \(\frac{H}{f} \times \frac{W}{f} \).

This feature map is then processed by the bottleneck module \(Bottleneck\) to yield the latent representation $\mathbf{X}$. This step imposes paradigm-specific constraints—such as quantization for discrete models or KL-regularization for continuous ones. Finally, the decoder \(Decoder\) reconstructs the image $\hat{\mathbf{I}}$ from these latents:
\begin{equation}
\mathbf{X} = Bottleneck(\mathbf{L}), \quad \hat{\mathbf{I}} = Decoder(\mathbf{X}).
\end{equation}

In practice, each latent token $x \in \mathbf{X}$ is encouraged to follow a structured distribution (\eg, discrete or Gaussian), which facilitates subsequent generative modeling by making the latent space easier to model and sample from.

\subsection{Benchmarking Discrete and Continuous Tokenizers}
The primary distinction between discrete and continuous tokenizers lies in the design of the bottleneck. Discrete tokenizers typically rely on codebook lookup with hard assignments to discretize latent features, whereas continuous tokenizers impose bottlenecks through dimensionality reduction combined with regularization losses. This fundamental difference makes direct comparison between the two paradigms nontrivial. In practice, discrete tokenizers are commonly characterized by their codebook size~\cite{esser2021taming,yu2021vector}, while continuous tokenizers are often compared based on the dimensionality of their latent representations~\cite{rombach2022high,li2024autoregressive}.

\begin{figure}[t!]
    \centering    \includegraphics[width=0.9\linewidth]{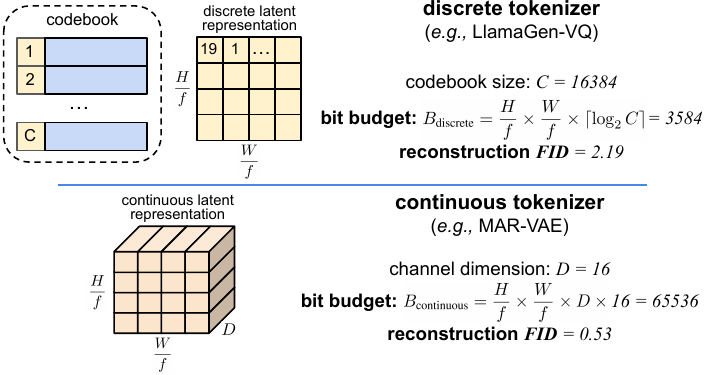}
    \vspace{-1ex}
    \caption{
    \textbf{A unified view for comparing discrete and continuous tokenizers.}
    By measuring information capacity in bits, we enable a direct comparison. The continuous tokenizer MAR-VAE~\cite{li2024autoregressive} outperforms the discrete tokenizer LlamaGen-VQ~\cite{sun2024autoregressive} in reconstruction quality, a result directly attributable to its substantially higher bit allocation.
    }
    \vspace{-2ex}
    \label{fig:bits_metric}
\end{figure}

To enable a unified and fair comparison across these paradigms, we evaluate both tokenizers using a common metric: the \textit{Bit Budget} ($B$). This metric quantifies the total information capacity allocated to the latent space, serving as a proxy for the nominal compression ratio.
Formally, consider an input image \(\mathbf{I}\) of height \(H\) and width \(W\), processed by a tokenizer with spatial downsampling factor \(f\). For a discrete tokenizer with codebook size \(C\), its bit budget is\footnote{We mainly discuss the most common single-scale and single-codebook tokenizers, whereas the formulation can be easily generalized to other cases such as multi-scale~\cite{tian2024visual} or multi-codebook~\cite{qu2025tokenflow,ma2025unitok}.}:
\begin{equation}
    B_{\text{discrete}} = \frac{H}{f} \times \frac{W}{f} \times \lceil \log_2 C \rceil.
\end{equation}
Conversely, for a continuous tokenizer with latent channel dimension \(D\), the bit budget is:
\begin{equation}
    B_{\text{continuous}} = \frac{H}{f} \times \frac{W}{f} \times 16D,
\end{equation}
where the constant factor \(16\) reflects mixed-precision training, with each latent channel represented using $16$ bits.
While bit budget $B$ defines the nominal capacity, the effective information content may be lower due to dead codebook entries or distributional regularization. This metric facilitates the direct comparison shown in \figref{fig:bits_metric}.

\begin{figure}[t!]
    \centering
    \includegraphics[width=0.85\linewidth]{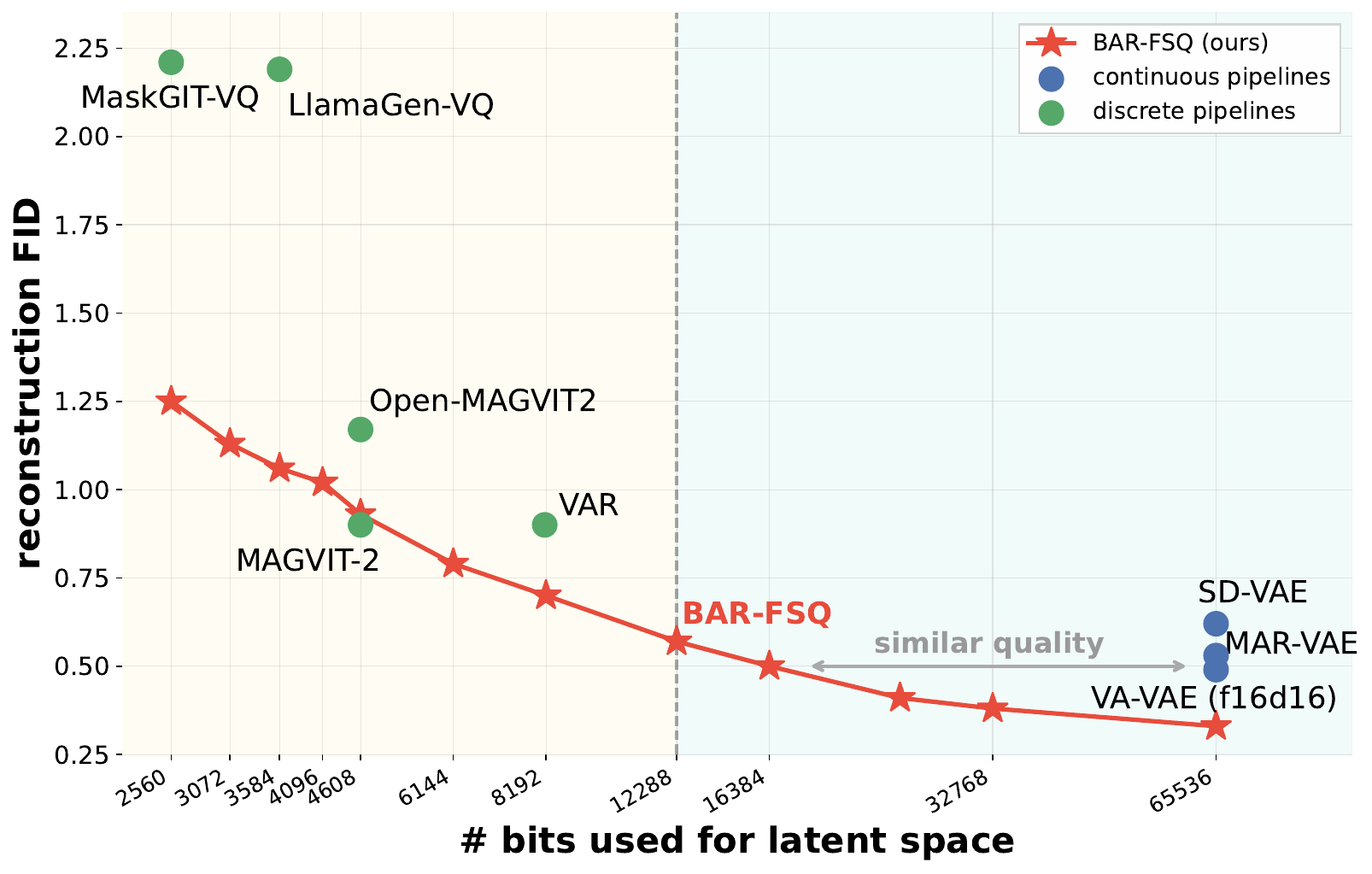}
    \caption{
    \textbf{Scaling \modelname's discrete tokenizer (\modelname-FSQ) with Bit Budget.} Standard discrete methods (green circles) historically lag behind continuous baselines (blue circles) primarily due to restricted bit allocation. By systematically scaling the codebook size, \modelname-FSQ (red curve) demonstrates that discrete tokenizer's reconstruction performance is not inherently bounded; it matches and further surpasses continuous reconstruction fidelity with increased bit budget, challenging the assumption that continuous latent spaces are required for high-fidelity reconstruction.
    }
    \vspace{-2ex}
    \label{fig:recon_comp}
\end{figure}

\begin{figure*}[t!]
    \centering
    \includegraphics[width=0.85\linewidth]{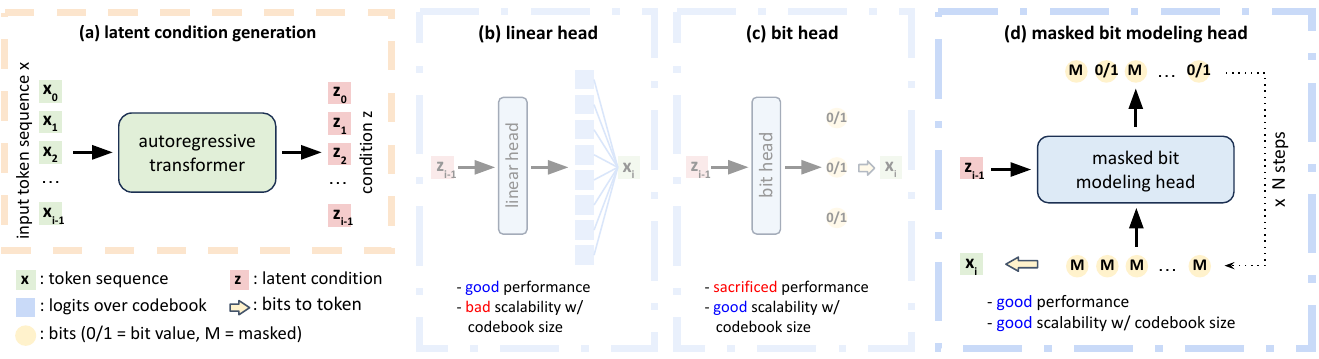}
    \caption{
    \textbf{Overview of the proposed \modelname framework.} We decompose autoregressive visual generation into two stages: context modeling and token prediction.
    (a) For context modeling, we employ an autoregressive transformer to generate latent conditions via causal attention.
    For the subsequent token prediction stage, we contrast our method with two baselines:
    (b) A standard linear head predicts logits over the full codebook. While effective for small vocabularies ($<2^{18}$), it fails to scale to larger sizes due to computational bottlenecks.
    (c) A bit-based head predicts bits directly; while scalable, it results in inferior generation quality.
    (d) The proposed Masked Bit Modeling (MBM) head generates bits via a progressive unmasking mechanism conditioned on the autoregressive transformer's output. Unlike the baselines, MBM achieves both exceptional scalability and superior generation quality.
    }
    \label{fig:generator}
\end{figure*}

\subsection{Discrete Tokenizers Beat Continuous Tokenizers}

Equipped with the Bit Budget metric, we conduct a systematic evaluation of existing discrete and continuous tokenizers. 
As shown in~\figref{fig:recon_comp}, we observe a distinct separation between the two paradigms: discrete tokenizers generally exhibit worse reconstruction quality while using substantially fewer bits. This discrepancy in compression ratio is non-negligible and can largely account for the inferior reconstruction performance observed in discrete methods.

Crucially, we identify a convergence trend: as we increase the number of bits allocated to the latent space, the performance of discrete tokenizers progressively improves, narrowing the gap with continuous tokenizers. This observation prompts a critical investigation: \textit{Is the perceived inferiority of discrete tokenizers intrinsic to the quantization bottleneck, or is it merely a consequence of insufficient bit allocation?}

To address this, we examine the effect of scaling the codebook size to approach the bit budget of continuous tokenizers. Since classical Vector Quantization (VQ) with learnable codebooks becomes computationally infeasible at extreme scales (\eg, $2^{256}$), we adopt the FSQ quantizer~\cite{mentzer2023finite}\footnote{The discussion here can easily generalize to other bit quantization such as LFQ~\cite{yu2023language} or BSQ~\cite{zhao2024image}.}. This allows us to scale smoothly without auxiliary quantization or entropy losses~\cite{chang2022maskgit,yu2023language,zhao2024image}.

For simplicity, we fix the number of latent tokens to $256$ (\ie, a downsampling ratio of $f=16$ for $256\times256$ images) and use $1$ bit per channel in the FSQ quantizer.
We then vary the latent channel dimension from $10$ to $12$, $14$, $16$, $18$, $32$, $64$, $128$, and $256$, corresponding to codebook sizes of $2^{10}$, $2^{12}$, $2^{14}$, $2^{16}$, $2^{18}$, $2^{32}$, $2^{64}$, $2^{128}$, and $2^{256}$, respectively.

The results, shown by the \modelname-FSQ curve in~\figref{fig:recon_comp}, demonstrate that reconstruction quality improves consistently with codebook size. Notably, when the bit budget increases beyond certain point, the discrete tokenizer achieves competitive or superior fidelity. For instance, at a budget of $65536$ bits, our discrete tokenizer attains an rFID of $0.33$, outperforming the SD-VAE (rFID $0.62$).

Furthermore, discrete tokenizers demonstrate superior efficiency in budget utilization. With only $16384$ bits, we achieve comparable performance (rFID $0.50$). This indicates that discrete method yields highly expressive representations even under strict constraints, leading to our core discovery:

\textit{The main performance bottleneck of discrete tokenizer lies in an insufficient bit budget, while scaling up codebook size enables discrete tokenization outperform continuous approaches.}

\subsection{Discrete Autoregressive Models Beat Diffusion}

While scaling the codebook size effectively resolves the reconstruction bottleneck (as established in the preceding subsection), it introduces a new, critical impediment to generative modeling: the \textit{vocabulary scaling problem}.

Standard autoregressive models face a prohibitive computational cliff as vocabularies expand. Projecting high-dimensional hidden states onto a vocabulary of millions ($2^{20}$) or billions ($2^{30}$) of entries renders the final linear prediction head intractable in terms of both memory and compute. Consequently, prior works typically cap codebook sizes at $2^{18}$ ($262144$), accepting a ceiling on reconstruction fidelity to preserve trainability. Furthermore, even when hardware permits, learning a reliable categorical distribution over such a vast space is statistically difficult, leading to a sharp degradation in generation quality~\cite{yu2023language}.

We empirically validate this limitation by training models with a standard \textit{linear} prediction head across different codebook sizes. The model works fine with limited codebook size but stops at $18$ bits (corresponding to vocabulary sizes $262144$); beyond this range, training becomes unaffordable under typical GPU memory constraints.
We also experimented with a \textit{bits-based} head that predicts the bit representation of target discrete token instead of the index over entire vocabulary~\cite{han2025infinity}. While this approach enables training with large codebook sizes, it consistently yields inferior performance across vocabulary sizes and suffers from severe degradation as the vocabulary scales.

\noindent \textbf{Prediction Head as a Bit Generator.} To overcome this, we disentangle the generator into two distinct functional components: an \textit{Autoregressive Transformer}, which captures global structure via causal attention, and a \textit{Prediction Head}, which projects latent embeddings onto specific discrete codes. This separation is critical: as codebook sizes scale, the autoregressive transformer remains computationally invariant; the entire burden of the exponential vocabulary growth is absorbed exclusively by the prediction head.

Unlike prior approaches relying on \textit{linear} or \textit{bit-based} projection, we propose a paradigm shift: rather than treating token prediction as a massive classification task, we formulate it as a conditional generation task. We introduce a \textbf{Masked Bit Modeling (MBM) Head}, which generates the target discrete token via an iterative, bit-wise unmasking process conditioned on the \textit{autoregressive transformer}'s output. The proposed prediction head is lightweight, typically requiring only a small number of additional forward passes to decode a discrete token.

\begin{figure}[t!]
    \centering
    \includegraphics[width=0.9\linewidth]{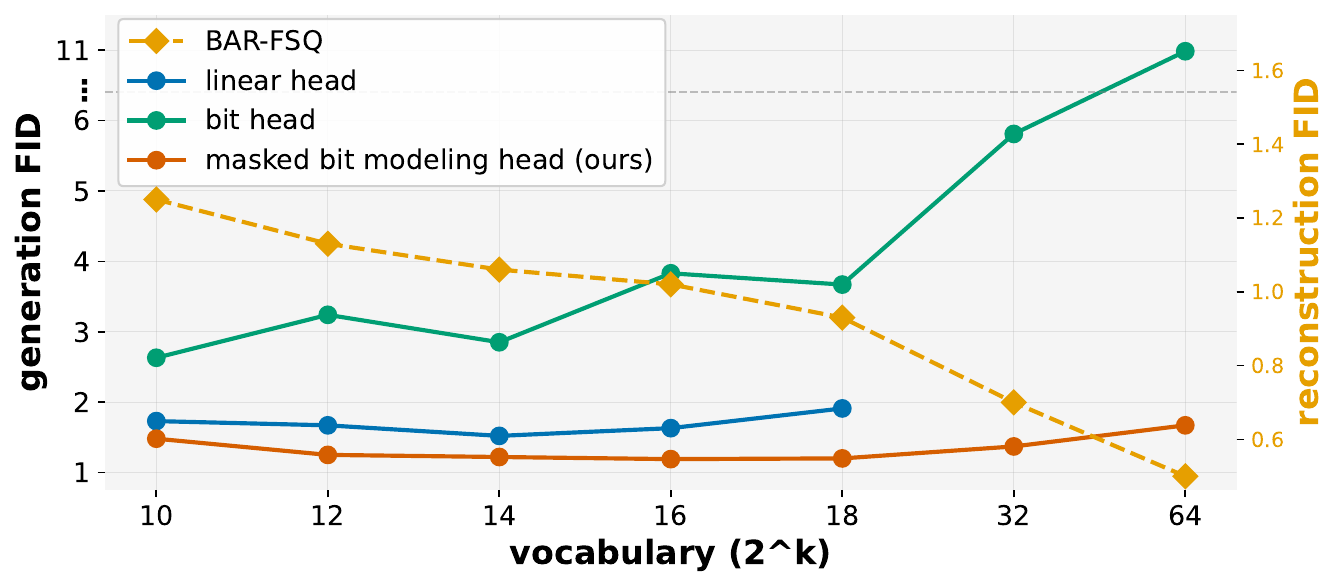}
    \caption{
    \textbf{Reconstruction and generation quality as a function of \modelname tokenizer's vocabulary size.} Unlike the linear head, the proposed \textit{masked bit modeling head} scales to arbitrary codebook sizes. Furthermore, it achieves a superior reconstruction--generation trade-off compared to the bit head.
    }
    \vspace{-3ex}
    \label{fig:rfid_gfid}
\end{figure}

\noindent \textbf{Formulation.}
Let $\mathcal{F}$ denote the autoregressive transformer~\cite{vaswani2017attention}.
Given a causal prefix of discrete tokens $\{x_1, x_2, \ldots, x_{i-1}\}$, where each token $x$ is represented by $k$-bit binary code, the autoregressive transformer maps the input to a sequence $\{z_1, z_2, \ldots, z_{i-1}\}$.
Specifically, for the prediction of the $i$-th token, we have:
\begin{equation}
    z_{i-1} = \mathcal{F}(\{x_1, x_2, \ldots, x_{i-1}\}),
\end{equation}

We utilize $z_{i-1}$ as a condition to predict the next token $x_i$ via a masked bit modeling head $\mathcal{G}$ parameterized by $\theta$:
\begin{equation}
    \hat{x}_i = \mathcal{G}_{\theta}\bigl(\text{Mask}_{bit}(x_i) \mid z_{i-1}, \mathcal{M}\bigr),
\end{equation}
where $\text{Mask}_{bit}(\cdot)$ randomly masks a subset of bits in $x_i$ by replacing them with a special mask token, and $\mathcal{M}$ denotes the masking ratio.

During training, we optimize cross-entropy loss between the predicted token $\hat{x}_i$ and the ground-truth token $x_i$:
\vspace{-1ex}
\begin{equation}
    \mathcal{L} = \frac{1}{n} \sum_{i=1}^{n} \text{CrossEntropy}_{bit}(x_i, \hat{x}_i),
\end{equation}
where $n$ is the sequence length, and $\text{CrossEntropy}_{bit}$ applies the loss in a bit-wise manner.

At inference, the next token is not selected via a single sampling step but is ``generated'' through a progressive bit-wise unmasking schedule~\cite{chang2022maskgit}.

As illustrated in~\figref{fig:generator}, this design offers two key advantages. First, in terms of \textbf{scalability}, decomposing the token into its constituent bits bypasses the need for a monolithic softmax over the entire vocabulary, reducing memory complexity from $\mathcal{O}(C)$ to $\mathcal{O}(\log_2 C)$, where $C=2^k$ is the codebook size. Second, regarding \textbf{robustness}, the bit-wise masking acts as a strong regularizer that consistently improves generation quality. As a result, the MBM head yields a superior trade-off between reconstruction (rFID) and generation (gFID) across all codebook scales, as shown in~\figref{fig:rfid_gfid}.

\noindent \textbf{Discussion.}
Unlike standard linear heads, which are constrained by fixed computational costs and memory requirements that scale linearly with vocabulary size, the proposed masked bit modeling head facilitates discrete generation with arbitrarily large vocabularies. Our results demonstrate consistent improvements over baselines, with the advantage becoming more pronounced at larger codebook sizes. This improved scaling behavior stems from the model's capacity to flexibly allocate more computation per token via progressive unmasking, a mechanism analogous to the iterative denoising process in diffusion models.
\vspace{-1ex}
\section{Experimental Results}

\subsection{Implementation Details}
\vspace{-0.5ex}
\noindent \textbf{Tokenizer.}
We build the discrete tokenizer using FSQ~\cite{mentzer2023finite}, incorporating several modern design choices from recent works~\cite{weber2024maskbit,tian2024visual,lu2025atoken,zheng2025diffusion} to better align with contemporary training recipes. Specifically, we initialize the encoder from SigLIP2-so400M~\cite{tschannen2025siglip} and apply an L2 loss against the original CLIP features to encourage semantic alignment in the latent space~\cite{zheng2025vision}. For the decoder, we use a ViT-L~\cite{dosovitskiy2020image} model trained from scratch, and we employ a frozen DINO model~\cite{caron2021emerging,sauer2023stylegan,tian2024visual,zheng2025diffusion} as the discriminator. The final training objective combines L1, L2, perceptual~\cite{zhang2018unreasonable}, Gram loss~\cite{lu2025atoken} and GAN losses~\cite{goodfellow2014generative}. The training is conducted for 40 epochs for ablation studies. For final models, we finetune the decoder for $40$ more epochs.

\vspace{-0.5ex}
\noindent \textbf{Generator.}
We build upon the state-of-the-art discrete autoregressive generation model RAR~\cite{yu2025randomized}. In addition, we augment the model with several architectural components commonly used in recent diffusion-based generators~\cite{yao2025reconstruction,li2025back,zheng2025diffusion}, including RoPE~\cite{su2024roformer}, SwiGLU~\cite{shazeer2020glu}, RMSNorm~\cite{zhang2019root}, and repeated class conditioning~\cite{li2025back}. The masked bit modeling head employs a 3-layer SwiGLU with adaLN~\cite{ba2016layer,peebles2023scalable}, which is lightweight and incurs only marginal extra cost. All training hyperparameters strictly follow the original RAR configuration. Training is conducted for $400$ epochs with a batch size of $2048$.

\vspace{-0.5ex}
\noindent \textbf{Sampling.}
We sample $50000$ images for FID computation using the evaluation code from~\cite{dhariwal2021diffusion}. When classifier-free guidance is employed, we adopt a simple linear guidance schedule~\cite{chang2023muse}.

\begin{table}[t]
\centering
\caption{\textbf{Scaling \modelname-FSQ codebook size ($C$) with different prediction heads.}
Unlike linear and bit-based baselines, the proposed \textit{masked bit modeling} (MBM) scales to arbitrary codebook sizes  while delivering superior generation quality.
}
\label{tab:pred_head}

\renewcommand{\arraystretch}{1.4}
\tablestyle{8.0pt}{1.0}

\adjustbox{max width=0.95\linewidth}{
\begin{tabular}{lcc|c|cc|cc}

\multirow[c]{3}{*}{\centering bits}
& \multirow[c]{3}{*}{\centering $C$}
& \multirow[c]{3}{*}{\centering rFID}
& \multirow[c]{3}{*}{\centering prediction head}
& \multicolumn{2}{c|}{\footnotesize w/o CFG}
& \multicolumn{2}{c}{\footnotesize w/ CFG} \\
& & & & \footnotesize gFID$\downarrow$ & \footnotesize IS$\uparrow$
       & \footnotesize gFID$\downarrow$ & \footnotesize IS$\uparrow$ \\
       
\shline

\multirow[c]{3}{*}{\centering 10}
& \multirow[c]{3}{*}{\centering 1024}
& \multirow[c]{3}{*}{\centering 1.25}
& \footnotesize linear   & \textbf{2.80} & 171.6 & 1.73 & 238.7  \\
& & & \footnotesize bit   & 10.77 & 107.6 & 2.63 & 213.9 \\
& &  & \footnotesize \baseline{\textbf{MBM}} & \baseline{3.10}    & \baseline{\textbf{180.6}} & \baseline{\textbf{1.48}}    & \baseline{\textbf{271.2}}  \\
\hline

\multirow[c]{3}{*}{\centering 12}
& \multirow[c]{3}{*}{\centering 4096}
& \multirow[c]{3}{*}{\centering 1.13}
& \footnotesize linear   & 2.70 & 180.3 & 1.67 & 248.9 \\
& & & \footnotesize bit   & 14.52 & 100.1 & 3.24 & 213.9  \\
& & & \footnotesize \baseline{\textbf{MBM}} & \baseline{\textbf{2.10}}  & \baseline{\textbf{207.9}} & \baseline{\textbf{1.25}}    & \baseline{\textbf{268.3}} \\
\hline

\multirow[c]{3}{*}{\centering 14}
& \multirow[c]{3}{*}{\centering 16384}
& \multirow[c]{3}{*}{\centering 1.06}

& \footnotesize linear   & 2.60 & 192.4 & 1.52 & 263.0 \\
& & & \footnotesize bit   & 14.57 & 96.9 & 2.85 & 224.5 \\
& & & \footnotesize \baseline{\textbf{MBM}} & \baseline{\textbf{1.71}}   & \baseline{\textbf{240.8}} & \baseline{\textbf{1.22}}    & \baseline{\textbf{292.2}}  \\
\hline

\multirow[c]{3}{*}{\centering 16}
& \multirow[c]{3}{*}{\centering 65536}
& \multirow[c]{3}{*}{\centering 1.02}
& \footnotesize linear   & 2.90 & 182.7 & 1.63 & 253.2 \\
& & & \footnotesize bit   & 23.11 & 71.8 & 3.83 & 215.0 \\
& & & \footnotesize \baseline{\textbf{MBM}} & \baseline{\textbf{1.68}}   & \baseline{\textbf{231.6}} & \baseline{\textbf{1.19}}    & \baseline{\textbf{282.3}} \\
\hline

\multirow[c]{3}{*}{\centering 18}
& \multirow[c]{3}{*}{\centering 262144}
& \multirow[c]{3}{*}{\centering 0.93}
& \footnotesize linear   & 3.45 & 172.9 & 1.91 & 241.0 \\
& & & \footnotesize bit   & 21.81 & 78.7 & 3.67 & 223.0 \\
& & & \footnotesize \baseline{\textbf{MBM}} & \baseline{\textbf{1.77}}    & \baseline{\textbf{228.5}} & \baseline{\textbf{1.20}}   & \baseline{\textbf{281.1}}  \\
\hline

\multirow[c]{3}{*}{\centering 32}
& \multirow[c]{3}{*}{\centering $\sim 4.29 \times 10^{9}$}
& \multirow[c]{3}{*}{\centering 0.70}
& \footnotesize linear   & OOM & OOM & OOM & OOM  \\
& & & \footnotesize bit   & 45.11 & 49.6 & 5.81 & 204.3 \\
&  & & \footnotesize \baseline{\textbf{MBM}} & \baseline{\textbf{2.37}}  & \baseline{\textbf{197.8}} & \baseline{\textbf{1.37}}   & \baseline{\textbf{292.1}}\\
\hline

\multirow[c]{3}{*}{\centering 64}
& \multirow[c]{3}{*}{\centering $\sim 1.84 \times 10^{19}$}
& \multirow[c]{3}{*}{\centering 0.50}
& \footnotesize linear   & OOM & OOM & OOM & OOM \\
&  & & \footnotesize bit   & 73.67 & 30.7 & 10.97 & 183.0 \\
&  & & \footnotesize \baseline{\textbf{MBM}} & \baseline{\textbf{2.60}}  & \baseline{\textbf{213.6}} & \baseline{\textbf{1.67}}   & \baseline{\textbf{295.0}}  \\

\end{tabular}
}
\vspace{-4ex}
\end{table}

\vspace{-1ex}
\subsection{Ablation Studies}

\begin{table*}[t]
\centering
\captionsetup{font=small}
\subcaptionsetup{font=footnotesize}
\caption{\textbf{Ablation studies on BAR design.} The rows labeled with \textcolor{gray}{gray} color indicate our choices for final models.}
\label{tab:ablation-2x2}
    
\setlength{\tabcolsep}{3pt}      
\renewcommand{\arraystretch}{0.95}
    \hspace{-4ex}
    \begin{subtable}[t]{0.48\textwidth}
        \centering
        \caption{\textbf{Impact of masking strategy during training.} \modelname demonstrates robustness across different masking strategies.}
        \label{tab:train_masking}
        
        \adjustbox{max width=\linewidth}{
\centering
\renewcommand{\arraystretch}{1.4}
\tablestyle{10.0pt}{1.1}
\begin{tabular}{l|c|cc|cc}

\multirow[c]{2}{*}{\centering bits}
& \multirow[c]{2}{*}{\centering masking strategy}
& \multicolumn{2}{c|}{\footnotesize w/o CFG}
& \multicolumn{2}{c}{\footnotesize w/ CFG} \\
& & \footnotesize FID$\downarrow$ & \footnotesize IS$\uparrow$
       & \footnotesize FID$\downarrow$ & \footnotesize IS$\uparrow$ \\
\shline

\multirow[c]{3}{*}{\centering 10}
& \footnotesize arccos   & 4.02 & 162.6 & 1.88 & 229.3 \\
& \footnotesize uniform   & 3.04 & 183.4 & 1.47 & 275.8 \\
& \footnotesize \baseline{logit-normal} & \baseline{3.10} & \baseline{180.6} & \baseline{1.48} & \baseline{271.2} \\
\hline

\multirow[c]{3}{*}{\centering 12}
& \footnotesize arccos   & 2.33 & 192.5 & 1.38 & 295.9 \\
& \footnotesize uniform   & 2.11 & 206.4 & 1.27 & 268.7 \\
& \footnotesize \baseline{logit-normal}  & \baseline{2.10} & \baseline{207.9} & \baseline{1.25} & \baseline{268.3} \\
\hline

\multirow[c]{3}{*}{\centering 16}
& \footnotesize arccos   & 1.89 & 218.1 & 1.46 & 311.8 \\
& \footnotesize uniform   & 1.78 & 225.4 & 1.22 & 293.7 \\

& \footnotesize \baseline{logit-normal}  & \baseline{1.68} & \baseline{231.6} & \baseline{1.19} & \baseline{282.3} \\
\hline

\multirow[c]{3}{*}{\centering 32}
& \footnotesize arccos   & 2.66 & 192.0 & 1.48 & 287.8 \\
& \footnotesize uniform   & 2.48 & 209.0 & 1.38 & 281.9 \\
& \footnotesize \baseline{logit-normal}  & \baseline{2.37} & \baseline{197.8} & \baseline{1.37} & \baseline{292.1} \\

\end{tabular}
\vspace{1ex}

          }
                
    \end{subtable}
    \hspace{4ex}
    \begin{subtable}[t]{0.38\textwidth}
        \centering
        \caption{\textbf{Scaling codebook size with different head sizes.}
        Increasing the head size improves performance, particularly for larger vocabularies. However, these benefits offer diminishing returns when Classifier-Free Guidance (CFG) is applied.
    }
    \label{tab:head_size}
        \adjustbox{max width=\linewidth}{
\centering
\renewcommand{\arraystretch}{1.4}
\tablestyle{10.0pt}{1.1}
\begin{tabular}{l|c|cc|cc}

\multirow[c]{2}{*}{\centering bits}
& \multirow[c]{1}{*}{\centering head size}
& \multicolumn{2}{c|}{\footnotesize w/o CFG}
& \multicolumn{2}{c}{\footnotesize w/ CFG} \\
& (\#layers $\times$ width) & \footnotesize FID$\downarrow$ & \footnotesize IS$\uparrow$
       & \footnotesize FID$\downarrow$ & \footnotesize IS$\uparrow$ \\
\shline

\multirow[c]{4}{*}{\centering 10}
& \footnotesize 3 $\times$ 1024 & 3.29    & 176.0 & 1.52    & 284.0 \\
& \footnotesize 3 $\times$ 1536 & 3.14    & 179.9 & 1.51    & 260.7 \\
& \baseline{\footnotesize 3 $\times$ 2048} & \baseline{3.10}    & \baseline{180.6} & \baseline{1.48} & \baseline{257.9} \\
& \footnotesize 6 $\times$ 2048 & 2.87  & 187.8 & 1.48 & 265.4 \\
\hline

\multirow[c]{4}{*}{\centering 12}
& \footnotesize 3 $\times$ 1024 & 2.16    & 206.2 & 1.26    & 286.3 \\
& \footnotesize 3 $\times$ 1536 & 2.07    & 195.3 & 1.25    & 280.1 \\
& \baseline{\footnotesize 3 $\times$ 2048} & \baseline{2.10}    & \baseline{207.9} & \baseline{1.25} & \baseline{268.3} \\
& \footnotesize 6 $\times$ 2048 & 2.02 & 198.2 & 1.27 & 279.5 \\
\hline

\multirow[c]{4}{*}{\centering 16}
& \footnotesize 3 $\times$ 1024 & 1.85    & 216.3 & 1.21    & 304.0 \\
& \footnotesize 3 $\times$ 1536 & 1.79    & 222.5 & 1.20    & 291.2 \\
& \footnotesize \baseline{3 $\times$ 2048} & \baseline{1.68}    & \baseline{231.6} & \baseline{1.19} & \baseline{282.3} \\
& \footnotesize 6 $\times$ 2048 & 1.63    & 233.8 & 1.19 & 293.2 \\
\hline

\multirow[c]{4}{*}{\centering 32}
& \footnotesize 3 $\times$ 1024 & 2.68    & 198.7 & 1.45    & 282.9 \\
& \footnotesize 3 $\times$ 1536 & 2.44    & 205.8 & 1.41    & 278.6 \\
& \baseline{\footnotesize 3 $\times$ 2048} & \baseline{2.37}    & \baseline{197.8} & \baseline{1.37} & \baseline{292.1} \\
& \footnotesize 6 $\times$ 2048 & 2.10 & 210.0 & 1.31 & 290.1 \\

\end{tabular}

  }
        
    \end{subtable}


    \hspace{3ex}
    \begin{subtable}[t]{0.45\textwidth}
        \centering
        \caption{\textbf{Impact of sampling strategy.} More steps advances results, while back-loading schedule further improves with CFG.}
        \label{tab:sampling}
        \adjustbox{max width=\linewidth}{
\centering
\renewcommand{\arraystretch}{1.4}
\tablestyle{10.0pt}{1.1}
\begin{tabular}{l|c|cc|cc}

\multirow[c]{2}{*}{\centering bits}
& \multirow[c]{2}{*}{\centering bits unmasking schedule}
& \multicolumn{2}{c|}{\footnotesize w/o CFG}
& \multicolumn{2}{c}{\footnotesize w/ CFG} \\
& & \footnotesize FID$\downarrow$ & \footnotesize IS$\uparrow$
       & \footnotesize FID$\downarrow$ & \footnotesize IS$\uparrow$ \\
\shline

\multirow[c]{6}{*}{\centering 16}

& \footnotesize [8, 8]   & 3.73 & 191.1 & 1.55 & 272.5 \\

& \footnotesize [5, 5, 6]  & 1.95 & 218.2 & 1.20 & 291.3 \\

& \baseline{\footnotesize [4, 4, 4, 4]}  & \baseline{1.68} & \baseline{231.6} & 1.19 & 282.3 \\

& \footnotesize [3, 3, 3, 3, 4]  & 1.64 & 235.8 & 1.23 & 301.1 \\

& \footnotesize [2, 2, 3, 3, 3, 3]  & 1.64 & 230.0 & 1.22 & 307.2 \\

& \baseline{\footnotesize [2, 2, 5, 7]}  & 1.81 & 214.2 & \baseline{1.15} & \baseline{289.2} \\

\end{tabular}
  }
    \end{subtable}
    \hfill
    \begin{subtable}[t]{0.48\textwidth}
        \centering
        \vspace{4ex}
        \caption{\textbf{Efficient BAR.} Sampling are based on uniform schedules with $4$ bits unmasking steps per token for all methods. BAR enables better accuracy-cost trade-off.}
        \label{tab:token_shuffle}
        \adjustbox{max width=\linewidth}{
\centering
\renewcommand{\arraystretch}{1.4}
\tablestyle{10.0pt}{1.1}
\begin{tabular}{ccc|cc|ccc}

\multirow[c]{3}{*}{\centering patch size} & \multirow[c]{3}{*}{\centering token}
& \multirow[c]{3}{*}{\centering bits per token}
& \multicolumn{2}{c|}{\footnotesize w/o CFG}
& \multicolumn{3}{c}{\footnotesize w/ CFG} \\
& & & \footnotesize FID$\downarrow$ & \footnotesize IS$\uparrow$
       & \footnotesize FID$\downarrow$ & \footnotesize IS$\uparrow$
       & images / sec \\
       
\shline

\baseline{BAR-B} & \baseline{256} & \baseline{16}  & \baseline{\textbf{1.68}}   & \baseline{\textbf{231.6}} & \baseline{\textbf{1.19}}    & \baseline{\textbf{282.3}} & \baseline{24.9} \\

BAR-B/2 & 64 & 64 & 2.24  & 217.3 & 1.35  & 293.4 & 150.3 \\

BAR-B/4 & 16 & 256  & 3.50  & 212.4 & 2.34    & 274.7 & \textbf{445.5} \\

\end{tabular}
  }
    \end{subtable}
\end{table*}

We study the impact of different designs based on BAR-B, supported by results both without and with classifier-free guidance (CFG)~\cite{ho2022classifier} for a comprehensive analysis of how different designs affect performance.

\vspace{-0.5ex}
\noindent \textbf{Different Prediction Heads.}
As shown in~\tabref{tab:pred_head}, the linear head performs reasonably well when the codebook size is small, but it does not scale to large codebook sizes: when the vocabulary reaches $2^{32}$, training is no longer feasible within a reasonable resource budget.
Although the bits head~\cite{han2025infinity} partially alleviates the computational bottleneck by making generation with large vocabularies affordable, its generation quality is significantly inferior. Without CFG, all bits-head variants yield gFID values $>10$, and even with CFG, performance remains poor with gFID $>2.6$, indicating substantial degradation in generation quality. Besides, its performance degrades as vocabulary scales.

In contrast, the proposed masked bit modeling head not only scales naturally to arbitrary codebook sizes, but also consistently yields superior generation performance. Even with a large codebook of size $2^{32}$, it achieves a gFID of $1.37$, approaching state-of-the-art performance.

\vspace{-0.5ex}
\noindent \textbf{Masking Ratio Sampling Strategy.}
We evaluate different masking ratio sampling strategies during training in~\tabref{tab:train_masking}, specifically comparing arccos~\cite{besnier2023pytorch}, uniform, and logit-normal sampling~\cite{esser2024scaling}. In contrast to typical Masked Image Modeling (MIM) generative models~\cite{chang2022maskgit,besnier2023pytorch,yu2024image,weber2024maskbit}, which often favor tail-heavy distributions (\eg, arccos), \modelname does not require such skewing. Instead, simple uniform sampling performs remarkably well. Overall, \modelname demonstrates robustness across all strategies, achieving competitive generation quality in each case. We adopt logit-normal sampling as the default, as it yields a slight performance advantage, particularly for larger codebook sizes.

\vspace{-0.5ex}
\noindent \textbf{Prediction Head Size.}
We summarize the impact of prediction head capacity in~\tabref{tab:head_size}, varying both the number of layers and the hidden width. We observe consistent improvements in generation quality without CFG as the prediction head capacity increases, while the gains become less pronounced when CFG is applied. Interestingly, the benefits of a larger prediction head are more substantial for larger codebook sizes, suggesting that predicting discrete tokens from a larger space is inherently more challenging and therefore benefits from a stronger generative prediction head.

\vspace{-0.5ex}
\noindent \textbf{Sampling Strategy.}
We ablate sampling strategies in~\tabref{tab:sampling} along two dimensions: the number of sampling steps and the bit unmasking schedule. Increasing the number of sampling steps from $2$ to $3$ yields a significant improvement in generation quality, while further increasing the steps to $5$ or $6$ provides only marginal gains. We also evaluate a back-loading bit unmasking schedule and find that it improves performance with CFG, but slightly degrades performance without CFG, where a uniform unmasking schedule remains preferable.

\begin{table*}[t]
\centering
\caption{\textbf{ImageNet-1K $256\times256$ generation results.}
We report metrics with and without classifier-free guidance. BAR only adopts a simple linear guidance schedule, with no need for auto-guidance~\cite{karras2024guiding} from an external model that is used by other state-of-the-art methods~\cite{zheng2025diffusion}.
}
\label{tab:imagenet_256}
\tablestyle{1.1pt}{1.05}
\newcolumntype{M}{>{\centering\arraybackslash}p{0.9cm}}
\adjustbox{max width=0.7\linewidth}{
\begin{tabular}{l|cc|MMMM|MMMM}
 & &
& \multicolumn{4}{c|}{generation@256 w/o guidance} 
& \multicolumn{4}{c}{generation@256 w/ guidance} \\
\cline{4-11}
method & epochs & \#params 
& FID$\downarrow$ & IS$\uparrow$ & Prec.$\uparrow$ & Rec.$\uparrow$
& FID$\downarrow$ & IS$\uparrow$ & Prec.$\uparrow$ & Rec.$\uparrow$ \\
\shline

\baseline{pixel space} & \baseline{} & \baseline{} 
& \baseline{} & \baseline{} & \baseline{} & \baseline{}
& \baseline{} & \baseline{} & \baseline{} & \baseline{} \\
\hline

ADM~\cite{dhariwal2021diffusion} & 350 & 554M 
& 10.94 & 101.0 & 0.69 & 0.63
& 3.94 & 215.8 & 0.83 & 0.53 \\

JiT~\cite{li2025back} & 600 & 2B 
& - & - & - & -
& 1.82 & 292.6 & 0.79 & 0.62 \\

SiD2~\cite{hoogeboom2024simpler} & 1280 & - 
& - & - & - & -
& 1.38 & - & - & - \\

\shline
\baseline{continuous tokens} & \baseline{} 
& \baseline{} & \baseline{} & \baseline{} & \baseline{}
& \baseline{} & \baseline{} & \baseline{} & \baseline{} \\
\hline

DiT~\cite{peebles2023scalable} & 1400 & 675M
& 9.62 & 121.5 & 0.67 & 0.67
& 2.27 & 278.2 & 0.83 & 0.57 \\

SiT~\cite{ma2024sit} & 1400 & 675M
& 8.61 & 131.7 & 0.68 & 0.67
& 2.06 & 270.3 & 0.82 & 0.59 \\

DiMR~\cite{liu2024alleviating} & 800 & 1.1B
& 3.56 & - & - & -
& 1.63 & 292.5 & 0.79 & 0.63 \\

FlowAR~\cite{ren2024flowar} & 400 & 1.9B
& - & - & - & -
& 1.65 & 296.5 & 0.83 & 0.60 \\

MDTv2~\cite{gao2023masked} & 1080 & 676M
& - & - & - & -
& 1.58 & 314.7 & 0.79 & 0.65 \\

MAR~\cite{li2024autoregressive} & 800 & 943M
& 2.35 & 227.8 & 0.79 & 0.62
& 1.55 & 303.7 & 0.81 & 0.62 \\

\hline

\multirow[c]{2}{*}{\centering VA-VAE~\cite{yao2025reconstruction}} & 80 & \multirow[c]{2}{*}{\centering 675M}
& 4.29 & - & - & -
& - & - & - & - \\
 & 800 & 
& 2.17 & 205.6 & 0.77 & 0.65
& 1.35 & 295.3 & 0.79 & 0.65 \\

\hline

\multirow[c]{2}{*}{\centering REPA~\cite{yu2024representation}} & 80 & \multirow[c]{2}{*}{\centering 675M}
& 7.90 & 122.6 & 0.70 & 0.65
& - & - & - & - \\
 & 800 & 
& 5.78 & 158.3 & 0.70 & 0.68
& 1.29 & 306.3 & 0.79 & 0.64 \\

\hline
\multirow[c]{2}{*}{\centering DDT~\cite{wang2025ddt}} & 80 & \multirow[c]{2}{*}{\centering 675M}
& 6.62 & 135.2 & 0.69 & 0.67
& 1.52 & 263.7 & 0.78 & 0.63 \\
 & 400 & 
& 6.27 & 154.7 & 0.68 & 0.69
& 1.26 & 310.6 & 0.79 & 0.65 \\
\hline

xAR~\cite{ren2025beyond} & 800 & 1.1B
& - & - & - & -
& 1.24 & 301.6 & 0.83 & 0.64 \\

\hline
\multirow[c]{2}{*}{\centering RAE~\cite{zheng2025diffusion}} & 80 & \multirow[c]{2}{*}{\centering 839M}
& 2.16 & 214.8 & 0.82 & 0.59
& - & - & - & - \\
& 800 & 
& 1.51 & 242.9 & 0.79 & 0.63
& 1.13 & 262.6 & 0.78 & 0.67 \\

\shline
\baseline{discrete tokens} & \baseline{} & \baseline{} 
& \baseline{} & \baseline{} & \baseline{} & \baseline{}
& \baseline{} & \baseline{} & \baseline{} & \baseline{} \\
\hline

MaskGIT~\cite{chang2022maskgit} & 300 & 177M
& 6.18 & 182.1 & - & -
& - & - & - & - \\

Open-MAGVIT2~\cite{luo2024open} & 350 & 1.5B
& - & - & - & -
& 2.33 & 271.8 & 0.84 & 0.54 \\

LlamaGen~\cite{sun2024autoregressive} & 300 & 3.1B
& 9.38 & 112.9 & 0.69 & 0.67
& 2.18 & 263.3 & 0.81 & 0.58 \\

TiTok~\cite{yu2024image} & 800 & 287M
& 4.44 & 168.2 & - & -
& 1.97 & 281.8 & - & - \\

VAR~\cite{tian2024visual} & 350 & 2.0B
& - & - & - & -
& 1.92 & 323.1 & 0.82 & 0.59 \\

MAGVIT-v2~\cite{yu2023language} & 1080 & 307M
& 3.65 & 200.5 & - & -
& 1.78 & 319.4 & - & - \\ 

MaskBit~\cite{weber2024maskbit} & 1080 & 305M
& - & - & - & -
& 1.52 & 328.6 & - & - \\

RAR~\cite{yu2025randomized} & 400 & 1.5B
& - & - & - & -
& 1.48 & 326.0 & 0.80 & 0.63 \\

\hline
BAR-B (ours) & 400 & 415M
& 1.64 & 230.4 & 0.80 & 0.62

& 1.13 & 289.0 & 0.77 & 0.66 \\

\hline
\multirow[c]{2}{*}{\centering BAR-L (ours)} & 80 & \multirow[c]{2}{*}{\centering 1.1B}
& \textbf{1.71} & 224.3 & 0.80 & 0.63
& \textbf{1.15} & 288.7 & 0.77 & 0.66 \\
& 400 & 
& \textbf{1.42} & 236.2 & 0.79 & 0.65

& \textbf{0.99} & 296.9 & 0.77 & 0.69 \\

\end{tabular}
}
\vspace{-2ex}
\end{table*}

\vspace{-0.5ex}
\noindent \textbf{Efficient Generation with Token-Shuffling.}
The proposed MBM head offers an additional advantage by enabling efficient visual generation trading off sequence length and bits per token using patch size, similar to prior practices~\cite{rombach2022high,peebles2023scalable,ma2025token}. By shuffling from tokens to bits (for example, flattening and concatenating the bits of neighboring tokens), the effective token sequence length can be significantly reduced, enabling more efficient generation. As shown in~\tabref{tab:token_shuffle}, BAR provides a flexible mechanism to trade off generation quality and computational cost by balancing computation between the autoregressive transformer and the masked bit modeling head. Specifically, BAR-B can downsample the latent space by $2\times$ (named BAR-B/2 with patch size $2$), resulting in $4\times$ fewer tokens, while incurring only a modest degradation in performance (from $1.68$ to $2.24$ without CFG and from $1.19$ to $1.35$ with CFG). Consequently, sampling throughput increases substantially, from $24.9$ images per second to $150.3$ images per second. More aggressive downsampling leads to BAR-B/4 (patch size $4$), which further improves the sampling speed to $445.5$ images per second.
\vspace{-1ex}
\subsection{Main Results}

We report BAR results against state-of-the-art methods on the ImageNet-1K benchmarks at resolutions $256 \times 256$. For all results reported, we use the official ADM scripts~\cite{dhariwal2021diffusion} to ensure a fair comparison.

\noindent \textbf{ImageNet 256$\times$256.}
We summarize the results in~\tabref{tab:imagenet_256}. We observe that BAR-B, despite having only $415$M parameters, achieves substantially better performance than prior state-of-the-art discrete generation methods. In particular, BAR-B uses only one quarter of the model size of RAR ($415$M \vs \ $1.5$B), yet attains significantly higher generation quality (gFID $1.13$ \vs \ $1.48$). It also outperforms other discrete approaches by a clear margin, including VAR ($1.13$ \vs \ $1.92$) and LlamaGen ($1.13$ \vs \ $2.18$).

Moreover, BAR-B already surpasses state-of-the-art diffusion models based on continuous pipelines. Specifically, BAR-B outperforms xAR, which is approximately $3\times$ larger in model size, achieving a gFID of $1.13$ compared to $1.24$. Despite its compact size, BAR-B exceeds the performance of several strong diffusion-based models, including DDT ($1.13$ \vs \  $1.26$), VA-VAE ($1.13$ \vs \  $1.35$), and MAR ($1.13$ \vs \  $1.55$). Compared to the concurrent work RAE, the two methods achieve comparable performance at gFID $1.13$.

Scaling BAR-B to a larger model yields BAR-L, which further improves performance and significantly outperforms all prior methods, both discrete and continuous, achieving a new state-of-the-art gFID of $0.99$. Notably, BAR-L not only sets a new record under classifier-free guidance ($0.99$ \vs \  $1.13$ for RAE), but also establishes a new best result without guidance ($1.42$ \vs \  $1.51$ for RAE).

\vspace{-0.5ex}
\noindent \textbf{Sampling Speed.}
\begin{table}[t]
\centering
\caption{\textbf{Sampling throughput (including de-tokenization process).} All are benchmarked using a single H200, with float32 precision. BAR only uses KV-cache without further optimization.
}
\label{tab:sampling_speed}

\tablestyle{3.0pt}{1.05}
\adjustbox{max width=0.85\linewidth}{
\begin{tabular}{l|ccc}
method  & \#params & FID$\downarrow$ & images / sec \\

\shline

PAR-4×~\cite{wang2025parallelized} & 3.1B & 2.29 & 4.92 \\
VAR~\cite{tian2024visual} & 2.0B & 1.92 & 8.08 \\
MeanFlow~\cite{geng2025mean} & 676M & 2.20 & 151.48 \\
\hline

BAR-B/4 (ours) & 416M & 2.34  & \textbf{445.48} \\

BAR-B/2 (ours) & 415M & 1.35  & 150.52 \\

\hline
MAR~\cite{li2024autoregressive} & 943M & 1.55  & 1.19 \\
VA-VAE~\cite{yao2025reconstruction} & 675M & 1.35  & 1.51 \\
DDT~\cite{wang2025ddt} & 675M & 1.26  & 1.62  \\
xAR~\cite{ren2025beyond} & 1.1B & 1.24  & 2.03  \\
RAE~\cite{zheng2025diffusion} & 839M & 1.13 & 6.62  \\
\hline
BAR-B (ours) & 415M & 1.13  & 24.33 \\
BAR-L (ours) & 1.1B & \textbf{0.99}  & 10.65 \\
\end{tabular}}
\vspace{-4ex}
\end{table}
We compare BAR with state-of-the-art methods in terms of sampling speed in~\tabref{tab:sampling_speed}. Notably, the \textbf{efficient} variants of BAR achieve an excellent trade-off between generation quality and sampling efficiency. BAR-B/2, with a gFID of $1.35$, not only outperforms all efficient generation methods such as PAR (gFID $2.29$) and VAR (gFID $1.92$), but also achieves substantially faster sampling speeds, with $30.59\times$ and $18.64\times$ speedups over PAR and VAR, respectively. Even when compared to single-step diffusion models such as MeanFlow, BAR-B/2 demonstrates superior generation quality (gFID $1.35$ \vs \  $2.20$) at comparable sampling speed ($150.52$ \vs \  $151.48$ images per second). The most efficient variant, BAR-B/4, achieves generation quality comparable to MeanFlow (gFID $2.34$ \vs \  $2.20$), while producing samples $2.94\times$ faster.

In more performance-oriented comparisons, BAR-B achieves state-of-the-art visual quality while being $20.45\times$, $16.11\times$, $15.02\times$, $11.99\times$, and $3.68\times$ faster than MAR, VA-VAE, DDT, xAR, and RAE, respectively. Notably, the best-performing variant BAR-L not only sets a new state-of-the-art record with gFID $0.99$, but also maintains a clear advantage in sampling speed, achieving $8.95\times$ speedup over MAR, $5.25\times$ over xAR, and $1.61\times$ over RAE.
\vspace{-2ex}
\section{Conclusion}
\vspace{-1ex}
In this paper, we presented a unified and fair comparison between discrete and continuous visual tokenizers. We showed that differences in compression ratio, as measured by the number of bits allocated to the latent space, constitute a dominant factor underlying the observed performance differences between discrete and continuous tokenizers. When operating under comparable bit budgets, discrete tokenizers can match or even outperform their continuous counterparts.

Building on this analysis, we introduced a novel generative prediction head that models discrete tokens by generating their bit representations. This design enables efficient and effective discrete generation with arbitrarily large vocabularies, overcoming a key limitation of prior discrete generative models. As a result, the proposed \emph{masked bit autoregressive modeling} framework establishes a new state of the art, substantially outperforming both existing discrete methods and strong continuous baselines.

\clearpage
\noindent \textbf{Impact Statement.}
This work advances the field of visual generation by demonstrating that discrete tokenizers can match or surpass continuous approaches when given sufficient information capacity, while achieving faster sampling speeds and more efficient training. By making high-quality image generation more computationally accessible through such methods, this research could democratize generative AI for researchers with limited resources and reduce the environmental impact of large-scale generation tasks. However, the improved quality and efficiency of these models also amplify concerns around potential misuse for creating deepfakes, spreading misinformation, or generating harmful content at scale. These advances underscore the critical importance of developing robust detection methods, implementing responsible access controls, and establishing clear ethical guidelines for deployment.
\bibliography{icml2026}

@String(CVPR= {IEEE Conf. Comput. Vis. Pattern Recog.})

@String(ICCV= {Int. Conf. Comput. Vis.})

@String(ECCV= {Eur. Conf. Comput. Vis.})

@String(NIPS= {Adv. Neural Inform. Process. Syst.})

@String(ICLR = {Int. Conf. Learn. Represent.})

@String(CVPR  = {CVPR})

@String(ICCV  = {ICCV})

@String(ECCV  = {ECCV})

@String(NIPS  = {NeurIPS})

@String(ICLR  = {ICLR})

@String(ICML = {ICML})

@String(TMLR = {TMLR})

@article{wang2023seggpt,
  title={Seggpt: Segmenting everything in context},
  author={Wang, Xinlong and Zhang, Xiaosong and Cao, Yue and Wang, Wen and Shen, Chunhua and Huang, Tiejun},
  journal={arXiv preprint arXiv:2304.03284},
  year={2023}
}

@article{cui2025emu3,
  title={Emu3. 5: Native multimodal models are world learners},
  author={Cui, Yufeng and Chen, Honghao and Deng, Haoge and Huang, Xu and Li, Xinghang and Liu, Jirong and Liu, Yang and Luo, Zhuoyan and Wang, Jinsheng and Wang, Wenxuan and others},
  journal={arXiv preprint arXiv:2510.26583},
  year={2025}
}

@article{deng2025emerging,
  title={Emerging properties in unified multimodal pretraining},
  author={Deng, Chaorui and Zhu, Deyao and Li, Kunchang and Gou, Chenhui and Li, Feng and Wang, Zeyu and Zhong, Shu and Yu, Weihao and Nie, Xiaonan and Song, Ziang and others},
  journal={arXiv preprint arXiv:2505.14683},
  year={2025}
}

@article{wiedemer2025video,
  title={Video models are zero-shot learners and reasoners},
  author={Wiedemer, Thadd{\"a}us and Li, Yuxuan and Vicol, Paul and Gu, Shixiang Shane and Matarese, Nick and Swersky, Kevin and Kim, Been and Jaini, Priyank and Geirhos, Robert},
  journal={arXiv preprint arXiv:2509.20328},
  year={2025}
}

@article{karras2024guiding,
  title={Guiding a diffusion model with a bad version of itself},
  author={Karras, Tero and Aittala, Miika and Kynk{\"a}{\"a}nniemi, Tuomas and Lehtinen, Jaakko and Aila, Timo and Laine, Samuli},
  journal=NIPS,
  year={2024}
}

@article{wang2025bridging,
  title={Bridging continuous and discrete tokens for autoregressive visual generation},
  author={Wang, Yuqing and Lin, Zhijie and Teng, Yao and Zhu, Yuanzhi and Ren, Shuhuai and Feng, Jiashi and Liu, Xihui},
  journal={arXiv preprint arXiv:2503.16430},
  year={2025}
}

@article{ma2025unitok,
  title={Unitok: A unified tokenizer for visual generation and understanding},
  author={Ma, Chuofan and Jiang, Yi and Wu, Junfeng and Yang, Jihan and Yu, Xin and Yuan, Zehuan and Peng, Bingyue and Qi, Xiaojuan},
  journal={arXiv preprint arXiv:2502.20321},
  year={2025}
}

@inproceedings{qu2025tokenflow,
  title={Tokenflow: Unified image tokenizer for multimodal understanding and generation},
  author={Qu, Liao and Zhang, Huichao and Liu, Yiheng and Wang, Xu and Jiang, Yi and Gao, Yiming and Ye, Hu and Du, Daniel K and Yuan, Zehuan and Wu, Xinglong},
  booktitle=CVPR,
  year={2025}
}

@article{lu2025atoken,
  title={Atoken: A unified tokenizer for vision},
  author={Lu, Jiasen and Song, Liangchen and Xu, Mingze and Ahn, Byeongjoo and Wang, Yanjun and Chen, Chen and Dehghan, Afshin and Yang, Yinfei},
  journal={arXiv preprint arXiv:2509.14476},
  year={2025}
}

@article{ma2025token,
  title={Token-shuffle: Towards high-resolution image generation with autoregressive models},
  author={Ma, Xu and Sun, Peize and Ma, Haoyu and Tang, Hao and Ma, Chih-Yao and Wang, Jialiang and Li, Kunpeng and Dai, Xiaoliang and Shi, Yujun and Ju, Xuan and others},
  journal={arXiv preprint arXiv:2504.17789},
  year={2025}
}

@inproceedings{sauer2023stylegan,
  title={Stylegan-t: Unlocking the power of gans for fast large-scale text-to-image synthesis},
  author={Sauer, Axel and Karras, Tero and Laine, Samuli and Geiger, Andreas and Aila, Timo},
  booktitle=ICML,
  year={2023}
}

@article{geng2025mean,
  title={Mean flows for one-step generative modeling},
  author={Geng, Zhengyang and Deng, Mingyang and Bai, Xingjian and Kolter, J Zico and He, Kaiming},
  journal={arXiv preprint arXiv:2505.13447},
  year={2025}
}

@inproceedings{wang2025parallelized,
  title={Parallelized autoregressive visual generation},
  author={Wang, Yuqing and Ren, Shuhuai and Lin, Zhijie and Han, Yujin and Guo, Haoyuan and Yang, Zhenheng and Zou, Difan and Feng, Jiashi and Liu, Xihui},
  booktitle=CVPR,
  year={2025}
}

@inproceedings{yao2025reconstruction,
  title={Reconstruction vs. generation: Taming optimization dilemma in latent diffusion models},
  author={Yao, Jingfeng and Yang, Bin and Wang, Xinggang},
  booktitle=CVPR,
  year={2025}
}

@article{yu2024representation,
  title={Representation alignment for generation: Training diffusion transformers is easier than you think},
  author={Yu, Sihyun and Kwak, Sangkyung and Jang, Huiwon and Jeong, Jongheon and Huang, Jonathan and Shin, Jinwoo and Xie, Saining},
  journal=ICLR,
  year={2025}
}

@inproceedings{esser2024scaling,
  title={Scaling rectified flow transformers for high-resolution image synthesis},
  author={Esser, Patrick and Kulal, Sumith and Blattmann, Andreas and Entezari, Rahim and M{\"u}ller, Jonas and Saini, Harry and Levi, Yam and Lorenz, Dominik and Sauer, Axel and Boesel, Frederic and others},
  booktitle={ICML},
  year={2024}
}

@inproceedings{han2025infinity,
  title={Infinity: Scaling bitwise autoregressive modeling for high-resolution image synthesis},
  author={Han, Jian and Liu, Jinlai and Jiang, Yi and Yan, Bin and Zhang, Yuqi and Yuan, Zehuan and Peng, Bingyue and Liu, Xiaobing},
  booktitle=CVPR,
  year={2025}
}

@article{sun2024autoregressive,
  title={Autoregressive Model Beats Diffusion: Llama for Scalable Image Generation},
  author={Sun, Peize and Jiang, Yi and Chen, Shoufa and Zhang, Shilong and Peng, Bingyue and Luo, Ping and Yuan, Zehuan},
  journal={arXiv preprint arXiv:2406.06525},
  year={2024}
}

@inproceedings{caron2021emerging,
  title={Emerging properties in self-supervised vision transformers},
  author={Caron, Mathilde and Touvron, Hugo and Misra, Ishan and J{\'e}gou, Herv{\'e} and Mairal, Julien and Bojanowski, Piotr and Joulin, Armand},
  booktitle=ICCV,
  year={2021}
}

@article{weber2024maskbit,
  title={MaskBit: Embedding-free Image Generation via Bit Tokens},
  author={Weber, Mark and Yu, Lijun and Yu, Qihang and Deng, Xueqing and Shen, Xiaohui and Cremers, Daniel and Chen, Liang-Chieh},
  journal=TMLR,
  year={2024}
}

@article{zhao2024image,
  title={Image and video tokenization with binary spherical quantization},
  author={Zhao, Yue and Xiong, Yuanjun and Kr{\"a}henb{\"u}hl, Philipp},
  journal=ICLR,
  year={2025}
}

@article{yu2024image,
  title={An Image is Worth 32 Tokens for Reconstruction and Generation},
  author={Yu, Qihang and Weber, Mark and Deng, Xueqing and Shen, Xiaohui and Cremers, Daniel and Chen, Liang-Chieh},
  journal=NIPS,
  year={2024}
}

@article{luo2024open,
  title={Open-MAGVIT2: An Open-Source Project Toward Democratizing Auto-regressive Visual Generation},
  author={Luo, Zhuoyan and Shi, Fengyuan and Ge, Yixiao and Yang, Yujiu and Wang, Limin and Shan, Ying},
  journal={arXiv preprint arXiv:2409.04410},
  year={2024}
}

@article{team2024chameleon,
  title={Chameleon: Mixed-modal early-fusion foundation models},
  author={Team, Chameleon},
  journal={arXiv preprint arXiv:2405.09818},
  year={2024}
}

@article{li2024autoregressive,
  title={Autoregressive Image Generation without Vector Quantization},
  author={Li, Tianhong and Tian, Yonglong and Li, He and Deng, Mingyang and He, Kaiming},
  journal=NIPS,
  year={2024}
}

@article{tian2024visual,
  title={Visual autoregressive modeling: Scalable image generation via next-scale prediction},
  author={Tian, Keyu and Jiang, Yi and Yuan, Zehuan and Peng, Bingyue and Wang, Liwei},
  journal=NIPS,
  year={2024}
}

@article{wang2025ddt,
  title={Ddt: Decoupled diffusion transformer},
  author={Wang, Shuai and Tian, Zhi and Huang, Weilin and Wang, Limin},
  journal={arXiv preprint arXiv:2504.05741},
  year={2025}
}

@article{liu2024alleviating,
  title={Alleviating Distortion in Image Generation via Multi-Resolution Diffusion Models},
  author={Liu, Qihao and Zeng, Zhanpeng and He, Ju and Yu, Qihang and Shen, Xiaohui and Chen, Liang-Chieh},
  journal=NIPS,
  year={2024}
}

@inproceedings{ma2024sit,
  title={Sit: Exploring flow and diffusion-based generative models with scalable interpolant transformers},
  author={Ma, Nanye and Goldstein, Mark and Albergo, Michael S and Boffi, Nicholas M and Vanden-Eijnden, Eric and Xie, Saining},
  booktitle=ECCV,
  year={2024}
}

@inproceedings{zhai2022scaling,
  title={Scaling vision transformers},
  author={Zhai, Xiaohua and Kolesnikov, Alexander and Houlsby, Neil and Beyer, Lucas},
  booktitle=CVPR,
  year={2022}
}

@inproceedings{kingma2013auto,
  title={Auto-encoding variational bayes},
  author={Kingma, Diederik P and Welling, Max},
  booktitle=ICLR,
  year={2014}
}

@article{goodfellow2014generative,
  title={Generative adversarial nets},
  author={Goodfellow, Ian and Pouget-Abadie, Jean and Mirza, Mehdi and Xu, Bing and Warde-Farley, David and Ozair, Sherjil and Courville, Aaron and Bengio, Yoshua},
  journal=NIPS,
  year={2014}
}

@inproceedings{zhang2018unreasonable,
  title={The unreasonable effectiveness of deep features as a perceptual metric},
  author={Zhang, Richard and Isola, Phillip and Efros, Alexei A and Shechtman, Eli and Wang, Oliver},
  booktitle=CVPR,
  year={2018}
}

@article{zhu2024scaling,
  title={Scaling the codebook size of VQ-GAN to 100,000 with a utilization rate of 99\%},
  author={Zhu, Lei and Wei, Fangyun and Lu, Yanye and Chen, Dong},
  journal=NIPS,
  year={2024}
}

@inproceedings{zheng2023online,
  title={Online clustered codebook},
  author={Zheng, Chuanxia and Vedaldi, Andrea},
  booktitle=ICCV,
  year={2023}
}

@inproceedings{yu2021vector,
  title={Vector-quantized image modeling with improved vqgan},
  author={Yu, Jiahui and Li, Xin and Koh, Jing Yu and Zhang, Han and Pang, Ruoming and Qin, James and Ku, Alexander and Xu, Yuanzhong and Baldridge, Jason and Wu, Yonghui},
  booktitle=ICLR,
  year={2022}
}

@inproceedings{chang2022maskgit,
  title={Maskgit: Masked generative image transformer},
  author={Chang, Huiwen and Zhang, Han and Jiang, Lu and Liu, Ce and Freeman, William T},
  booktitle=CVPR,
  year={2022}
}

@inproceedings{mentzer2023finite,
  title={Finite scalar quantization: Vq-vae made simple},
  author={Mentzer, Fabian and Minnen, David and Agustsson, Eirikur and Tschannen, Michael},
  booktitle={ICLR},
  year={2024}
}

@inproceedings{esser2021taming,
  title={Taming transformers for high-resolution image synthesis},
  author={Esser, Patrick and Rombach, Robin and Ommer, Bjorn},
  booktitle=CVPR,
  year={2021}
}

@inproceedings{chang2023muse,
  title={Muse: Text-to-image generation via masked generative transformers},
  author={Chang, Huiwen and Zhang, Han and Barber, Jarred and Maschinot, AJ and Lezama, Jose and Jiang, Lu and Yang, Ming-Hsuan and Murphy, Kevin and Freeman, William T and Rubinstein, Michael and others},
  booktitle=ICML,
  year={2023}
}

@inproceedings{yu2023language,
  title={Language Model Beats Diffusion--Tokenizer is Key to Visual Generation},
  author={Yu, Lijun and Lezama, Jos{\'e} and Gundavarapu, Nitesh B and Versari, Luca and Sohn, Kihyuk and Minnen, David and Cheng, Yong and Gupta, Agrim and Gu, Xiuye and Hauptmann, Alexander G and others},
  booktitle=ICLR,
  year={2024}
}

@article{ho2020denoising,
  title={Denoising diffusion probabilistic models},
  author={Ho, Jonathan and Jain, Ajay and Abbeel, Pieter},
  journal=NIPS,
  year={2020}
}

@article{dhariwal2021diffusion,
  title={Diffusion models beat gans on image synthesis},
  author={Dhariwal, Prafulla and Nichol, Alexander},
  journal=NIPS,
  year={2021}
}

@inproceedings{sohl2015deep,
  title={Deep unsupervised learning using nonequilibrium thermodynamics},
  author={Sohl-Dickstein, Jascha and Weiss, Eric and Maheswaranathan, Niru and Ganguli, Surya},
  booktitle={ICML},
  year={2015}
}

@article{song2019generative,
  title={Generative modeling by estimating gradients of the data distribution},
  author={Song, Yang and Ermon, Stefano},
  journal=NIPS,
  year={2019}
}

@article{ba2016layer,
  title={Layer normalization},
  author={Ba, Jimmy Lei and Kiros, Jamie Ryan and Hinton, Geoffrey E},
  journal={arXiv preprint arXiv:1607.06450},
  year={2016}
}

@article{hoogeboom2024simpler,
  title={Simpler diffusion (sid2): 1.5 fid on imagenet512 with pixel-space diffusion},
  author={Hoogeboom, Emiel and Mensink, Thomas and Heek, Jonathan and Lamerigts, Kay and Gao, Ruiqi and Salimans, Tim},
  journal={arXiv preprint arXiv:2410.19324},
  year={2024}
}

@article{li2025fractal,
  title={Fractal generative models},
  author={Li, Tianhong and Sun, Qinyi and Fan, Lijie and He, Kaiming},
  journal={arXiv preprint arXiv:2502.17437},
  year={2025}
}

@article{zheng2025diffusion,
  title={Diffusion transformers with representation autoencoders},
  author={Zheng, Boyang and Ma, Nanye and Tong, Shengbang and Xie, Saining},
  journal={arXiv preprint arXiv:2510.11690},
  year={2025}
}

@inproceedings{rombach2022high,
  title={High-resolution image synthesis with latent diffusion models},
  author={Rombach, Robin and Blattmann, Andreas and Lorenz, Dominik and Esser, Patrick and Ommer, Bj{\"o}rn},
  booktitle=CVPR,
  year={2022}
}

@inproceedings{peebles2023scalable,
  title={Scalable diffusion models with transformers},
  author={Peebles, William and Xie, Saining},
  booktitle=ICCV,
  year={2023}
}

@inproceedings{gao2023masked,
  title={Masked diffusion transformer is a strong image synthesizer},
  author={Gao, Shanghua and Zhou, Pan and Cheng, Ming-Ming and Yan, Shuicheng},
  booktitle=ICCV,
  year={2023}
}

@article{ho2022classifier,
  title={Classifier-free diffusion guidance},
  author={Ho, Jonathan and Salimans, Tim},
  journal={arXiv preprint arXiv:2207.12598},
  year={2022}
}

@article{besnier2023pytorch,
  title={A Pytorch Reproduction of Masked Generative Image Transformer},
  author={Besnier, Victor and Chen, Mickael},
  journal={arXiv preprint arXiv:2310.14400},
  year={2023}
}

@article{ren2024flowar,
  title={Flowar: Scale-wise autoregressive image generation meets flow matching},
  author={Ren, Sucheng and Yu, Qihang and He, Ju and Shen, Xiaohui and Yuille, Alan and Chen, Liang-Chieh},
  journal={arXiv preprint arXiv:2412.15205},
  year={2024}
}

@article{lipman2022flow,
  title={Flow matching for generative modeling},
  author={Lipman, Yaron and Chen, Ricky TQ and Ben-Hamu, Heli and Nickel, Maximilian and Le, Matt},
  journal={arXiv preprint arXiv:2210.02747},
  year={2022}
}

@article{ren2025beyond,
  title={Beyond next-token: Next-x prediction for autoregressive visual generation},
  author={Ren, Sucheng and Yu, Qihang and He, Ju and Shen, Xiaohui and Yuille, Alan and Chen, Liang-Chieh},
  journal=ICCV,
  year={2025}
}

@article{agarwal2025cosmos,
  title={Cosmos world foundation model platform for physical ai},
  author={Agarwal, Niket and Ali, Arslan and Bala, Maciej and Balaji, Yogesh and Barker, Erik and Cai, Tiffany and Chattopadhyay, Prithvijit and Chen, Yongxin and Cui, Yin and Ding, Yifan and others},
  journal={arXiv preprint arXiv:2501.03575},
  year={2025}
}

@article{zhang2019root,
  title={Root mean square layer normalization},
  author={Zhang, Biao and Sennrich, Rico},
  journal=NIPS,
  year={2019}
}

@article{su2024roformer,
  title={Roformer: Enhanced transformer with rotary position embedding},
  author={Su, Jianlin and Ahmed, Murtadha and Lu, Yu and Pan, Shengfeng and Bo, Wen and Liu, Yunfeng},
  journal={Neurocomputing},
  year={2024}
}

@article{shazeer2020glu,
  title={Glu variants improve transformer},
  author={Shazeer, Noam},
  journal={arXiv preprint arXiv:2002.05202},
  year={2020}
}

@article{li2025back,
  title={Back to basics: Let denoising generative models denoise},
  author={Li, Tianhong and He, Kaiming},
  journal={arXiv preprint arXiv:2511.13720},
  year={2025}
}

@inproceedings{yu2025randomized,
  title={Randomized autoregressive visual generation},
  author={Yu, Qihang and He, Ju and Deng, Xueqing and Shen, Xiaohui and Chen, Liang-Chieh},
  booktitle=ICCV,
  year={2025}
}

@article{zheng2025vision,
  title={Vision foundation models as effective visual tokenizers for autoregressive image generation},
  author={Zheng, Anlin and Wen, Xin and Zhang, Xuanyang and Ma, Chuofan and Wang, Tiancai and Yu, Gang and Zhang, Xiangyu and Qi, Xiaojuan},
  journal=NIPS,
  year={2025}
}

@article{tschannen2025siglip,
  title={Siglip 2: Multilingual vision-language encoders with improved semantic understanding, localization, and dense features},
  author={Tschannen, Michael and Gritsenko, Alexey and Wang, Xiao and Naeem, Muhammad Ferjad and Alabdulmohsin, Ibrahim and Parthasarathy, Nikhil and Evans, Talfan and Beyer, Lucas and Xia, Ye and Mustafa, Basil and others},
  journal={arXiv preprint arXiv:2502.14786},
  year={2025}
}

@inproceedings{dosovitskiy2020image,
  title={An image is worth 16x16 words: Transformers for image recognition at scale},
  author={Dosovitskiy, Alexey and Beyer, Lucas and Kolesnikov, Alexander and Weissenborn, Dirk and Zhai, Xiaohua and Unterthiner, Thomas and Dehghani, Mostafa and Minderer, Matthias and Heigold, Georg and Gelly, Sylvain and Uszkoreit, Jakob and Houlsby, Neil},
  booktitle=ICLR,
  year={2021}
}

@article{vaswani2017attention,
  title={Attention is all you need},
  author={Vaswani, Ashish and Shazeer, Noam and Parmar, Niki and Uszkoreit, Jakob and Jones, Llion and Gomez, Aidan N and Kaiser, {\L}ukasz and Polosukhin, Illia},
  journal=NIPS,
  year={2017}
}

@inproceedings{bao2023all,
  title={All are worth words: A vit backbone for diffusion models},
  author={Bao, Fan and Nie, Shen and Xue, Kaiwen and Cao, Yue and Li, Chongxuan and Su, Hang and Zhu, Jun},
  booktitle=CVPR,
  year={2023}
}

@inproceedings{deng2009imagenet,
  title={Imagenet: A large-scale hierarchical image database},
  author={Deng, Jia and Dong, Wei and Socher, Richard and Li, Li-Jia and Li, Kai and Fei-Fei, Li},
  booktitle=CVPR,
  year={2009}
}

@article{hinton2006reducing,
  title={Reducing the dimensionality of data with neural networks},
  author={Hinton, Geoffrey E and Salakhutdinov, Ruslan R},
  journal={science},
  year={2006},
}

@inproceedings{liu2025flowing,
  title={Flowing from Words to Pixels: A Noise-Free Framework for Cross-Modality Evolution},
  author={Liu, Qihao and Yin, Xi and Yuille, Alan and Brown, Andrew and Singh, Mannat},
  booktitle={Proceedings of the Computer Vision and Pattern Recognition Conference},
  pages={2755--2765},
  year={2025}
}

@article{he2025flowtok,
  title={Flowtok: Flowing seamlessly across text and image tokens},
  author={He, Ju and Yu, Qihang and Liu, Qihao and Chen, Liang-Chieh},
  journal={arXiv preprint arXiv:2503.10772},
  year={2025}
}
\bibliographystyle{icml2026}

\clearpage
\newpage
\appendix


\section{Appendix}
\label{sec:appendix}

The supplementary material includes the following additional information:

\begin{itemize}
    \item \secref{sec:sup_hyper} provides the detailed hyper-parameters for the final BAR-FSQ and BAR models.
    \item \secref{sec:more_results} provides additional experimental results including BAR's results on the ImageNet-512 benchmark.
    \item \secref{sec:sup_vis_samples} provides more visualization samples of BAR models.
\end{itemize}

\section{Hyper-parameters for Final BAR Models}
\label{sec:sup_hyper}
\begin{table}[h!]
\centering
\caption{\textbf{Architecture configurations of BAR.} We follow prior works scaling up ViT~\cite{dosovitskiy2020image,zhai2022scaling} for different configurations.
}
\label{tab:arch}
\tablestyle{5.0pt}{1.05}

\begin{tabular}{l|ccccc}
model & depth & width & mlp & heads & \#params  \\
\shline
BAR-B & 24 & 768 & 3072 & 12  & 415M \\
BAR-L & 32 & 1280 & 5120 & 16  & 1108M  \\

\end{tabular}
\end{table}

\begin{table}[h!]
\centering
\caption{\textbf{Detailed hyper-parameters for final BAR-FSQ models.}
}
\label{tab:hparams_fsq}

\tablestyle{5.0pt}{1.1}
\adjustbox{max width=0.6\linewidth}{
\begin{tabular}{l|c}
config \quad\quad\quad\quad\quad\quad\quad\quad & value \\
\shline
\multicolumn{2}{c}{\textit{training hyper-params}} \\
\hline
optimizer & AdamW \\
learning rate & 1e-4 \\
weight decay & 0.0 \\
optimizer momentum & (0.9, 0.999) \\
total batch size & 256 \\
learning rate schedule & cosine decay \\
ending learning rate & 1e-5 \\
total epochs & 40 \\
warmup epochs & 2 \\
precision & bfloat16 \\
max grad norm & 1.0 \\
perceptual loss weight & 1.0 \\
clip l2 loss weight & 0.7 \\
gram loss weight & 10.0 \\
discriminator loss weight & 0.05 \\
discriminator kicks in epoch & 20 \\
reconstruction loss weight & 1.0 \\
training GPUs & 8 H200 \\
training time & 13 hrs \\

\end{tabular}}
\end{table}

\begin{table}[h!]
\centering
\caption{\textbf{Detailed hyper-parameters for final BAR models.}
}
\label{tab:hparams}

\tablestyle{5.0pt}{1.1}
\adjustbox{max width=0.6\linewidth}{
\begin{tabular}{l|c}
config \quad\quad\quad\quad\quad\quad\quad\quad & value \\
\shline
\multicolumn{2}{c}{\textit{training hyper-params}} \\
\hline
optimizer & AdamW \\
learning rate & 4e-4 \\
weight decay & 0.03 \\
optimizer momentum & (0.9, 0.96) \\
total batch size & 2048 \\
learning rate schedule & cosine decay \\
ending learning rate & 1e-5 \\
total epochs & 400 \\
warmup epochs & 100 \\
annealing start epoch & 200 \\
annealing end epoch & 300 \\
precision & bfloat16 \\
max grad norm & 1.0 \\
dropout rate & 0.0 (B) / 0.2 (L) \\
attn dropout rate & 0.0 (B) / 0.2 (L) \\
class label dropout rate & 0.1 \\
training GPUs & 16 H200 (B) / 32 H200 (L) \\
training time & 20 hrs (B) / 32 hrs (L) \\
\hline
\multicolumn{2}{c}{\textit{sampling hyper-params w/ CFG}}  \\
\hline
guidance schedule & linear \\
temperature & 2.5 (B) / 3.0 (L) \\
guidance scale & 5.0 (B) / 5.3 (L) \\
bits unmasking schedule & [2,2,5,7] \\
\hline
\multicolumn{2}{c}{\textit{sampling hyper-params w/o CFG}}  \\
\hline
temperature & 2.0 (B) / 2.4 (L) \\
bits unmasking schedule & [4,4,4,4] \\

\end{tabular}}
\end{table}

The BAR model configuration is detailed in~\tabref{tab:arch}.

We list the detailed training hyper-parameters and sampling hyper-parameters for all BAR-FSQ, BAR models in~\tabref{tab:hparams_fsq} and \tabref{tab:hparams}, resptively.

\section{More Experimental Results}
\label{sec:more_results}
\begin{table}[h!]
\centering
\caption{\footnotesize \textbf{ImageNet-1K $512\times 512$ generation results.} We report metrics with classifier-free guidance. \modelname only adopts a simple linear guidance schedule, with no need for auto-guidance~\cite{karras2024guiding} from an external model that is used by other state-of-the-art methods~\cite{zheng2025diffusion}. Due to computational constraints, we only train the model for $200$ epochs.
}
\label{tab:imagenet_512}

\tablestyle{1.0pt}{1.05}
\adjustbox{max width=0.75\linewidth}{
\begin{tabular}{l|ccc}
method & \#params & FID$\downarrow$ & IS$\uparrow$ \\
\shline
VQGAN~\cite{esser2021taming}&227M &26.52& 66.8\\
MaskGiT~\cite{chang2022maskgit}& 227M&7.32& 156.0\\
DiT~\cite{peebles2023scalable} &675M &3.04& 240.8 \\
DiMR~\cite{liu2024alleviating}& 525M&2.89 &289.8 \\
VAR~\cite{tian2024visual}  & 2.3B& 2.63 & 303.2\\
REPA~\cite{yu2024representation}&675M &2.08& 274.6 \\
xAR~\cite{ren2025beyond} & 608M & 1.70 & 281.5 \\
RAR~\cite{yu2025randomized} & 1.5B & 1.66 & 295.7 \\
DDT~\cite{wang2025ddt} & 675M & 1.28 & 305.1 \\
RAE~\cite{zheng2025diffusion} & 839M & 1.13 & 259.6 \\
\hline
BAR-L (ours) & 1.1B & \textbf{1.09} &  \textbf{311.1} \\
\end{tabular}}

\end{table}
We provide additional experimental results on ImageNet-512 in~\tabref{tab:imagenet_512}, where BAR demonstrates clear advantages over other methods.

\section{Visualization on Generated Samples}
\label{sec:sup_vis_samples}
\begin{figure*}[t]
    \centering
    \vspace{-24ex}
    \includegraphics[width=0.8\linewidth]{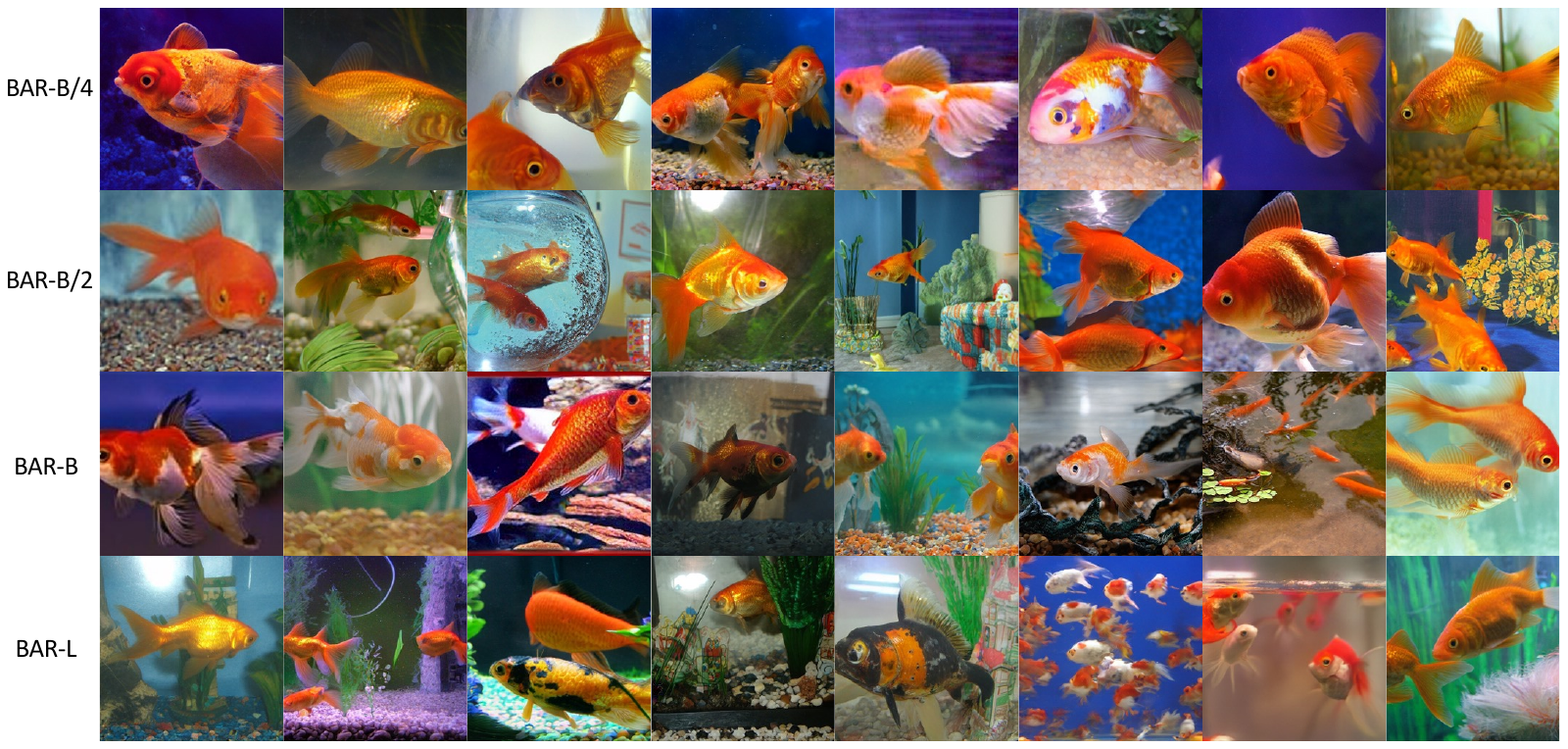}
    \vspace{-14ex}
    \caption{
    \textbf{Visualization samples from BAR models.} BAR is capable of generating high-fidelity image samples with great diversity. class idx 1: ``goldfish, Carassius auratus''.
    }
    \vspace{-37ex}
    \label{fig:vis_more1}
\end{figure*}

\begin{figure*}[t]
    \centering
    \includegraphics[width=0.8\linewidth]{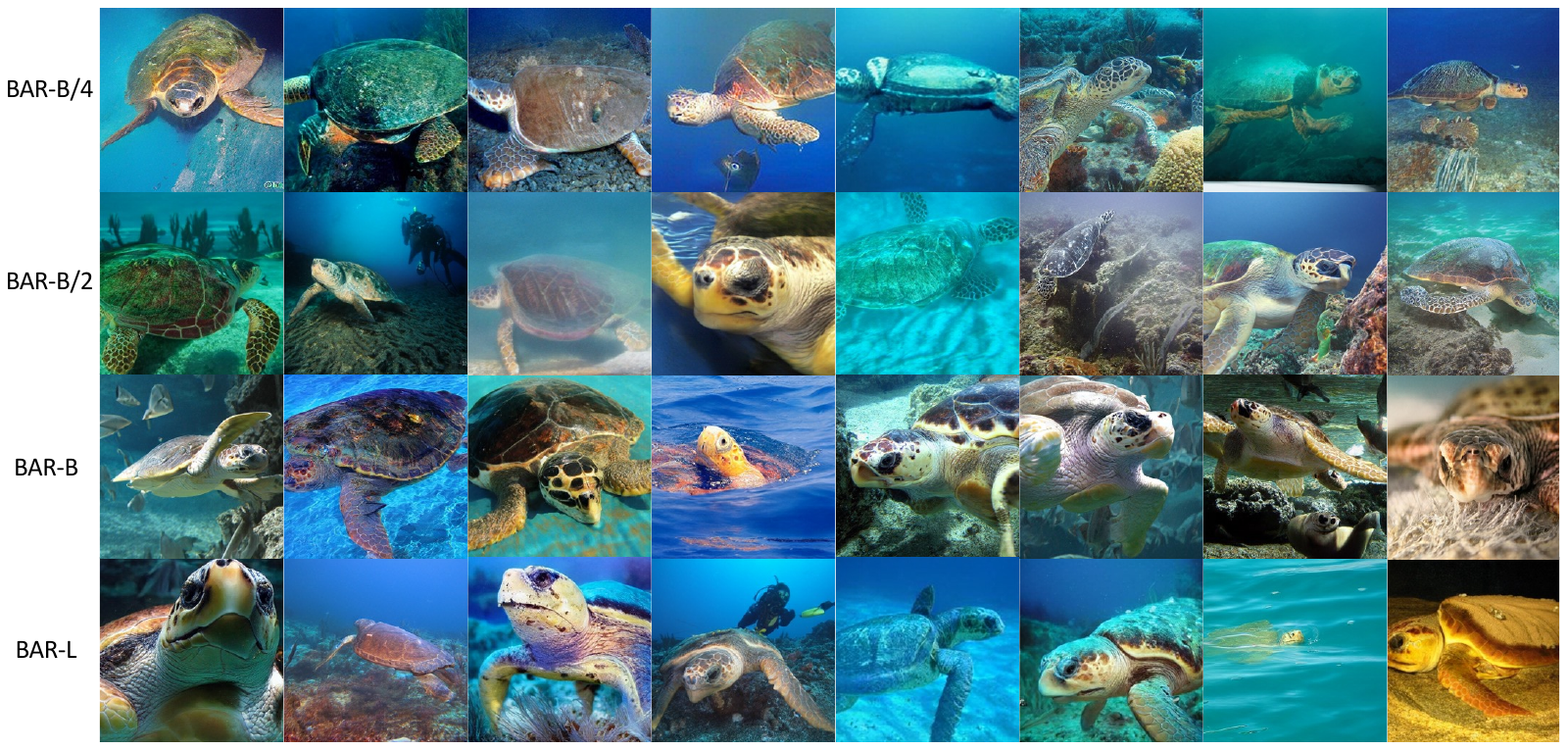}
    \vspace{-14ex}
    \caption{
    \textbf{Visualization samples from BAR models.} BAR is capable of generating high-fidelity image samples with great diversity. class idx 33: ``loggerhead, loggerhead turtle, Caretta caretta''.
    }
    \vspace{-37ex}
    \label{fig:vis_more2}
\end{figure*}

\begin{figure*}[t]
    \centering
    \includegraphics[width=0.8\linewidth]{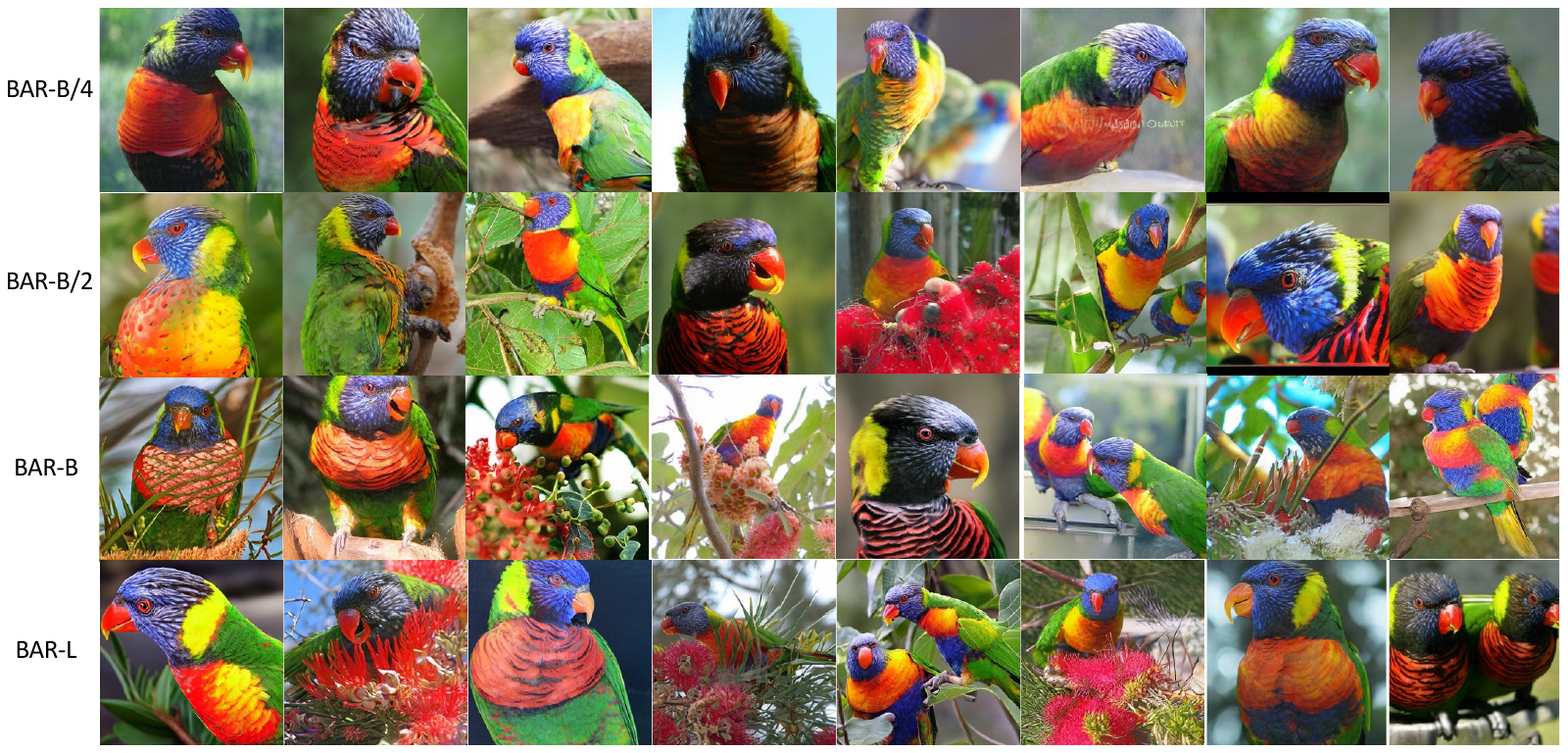}
    \vspace{-14ex}
    \caption{
    \textbf{Visualization samples from BAR models.} BAR is capable of generating high-fidelity image samples with great diversity. class idx 90: ``lorikeet''.
    }
    \vspace{-15ex}
    \label{fig:vis_more3}
\end{figure*}

\begin{figure*}[t]
    \centering
    \vspace{-24ex}
    \includegraphics[width=0.8\linewidth]
    {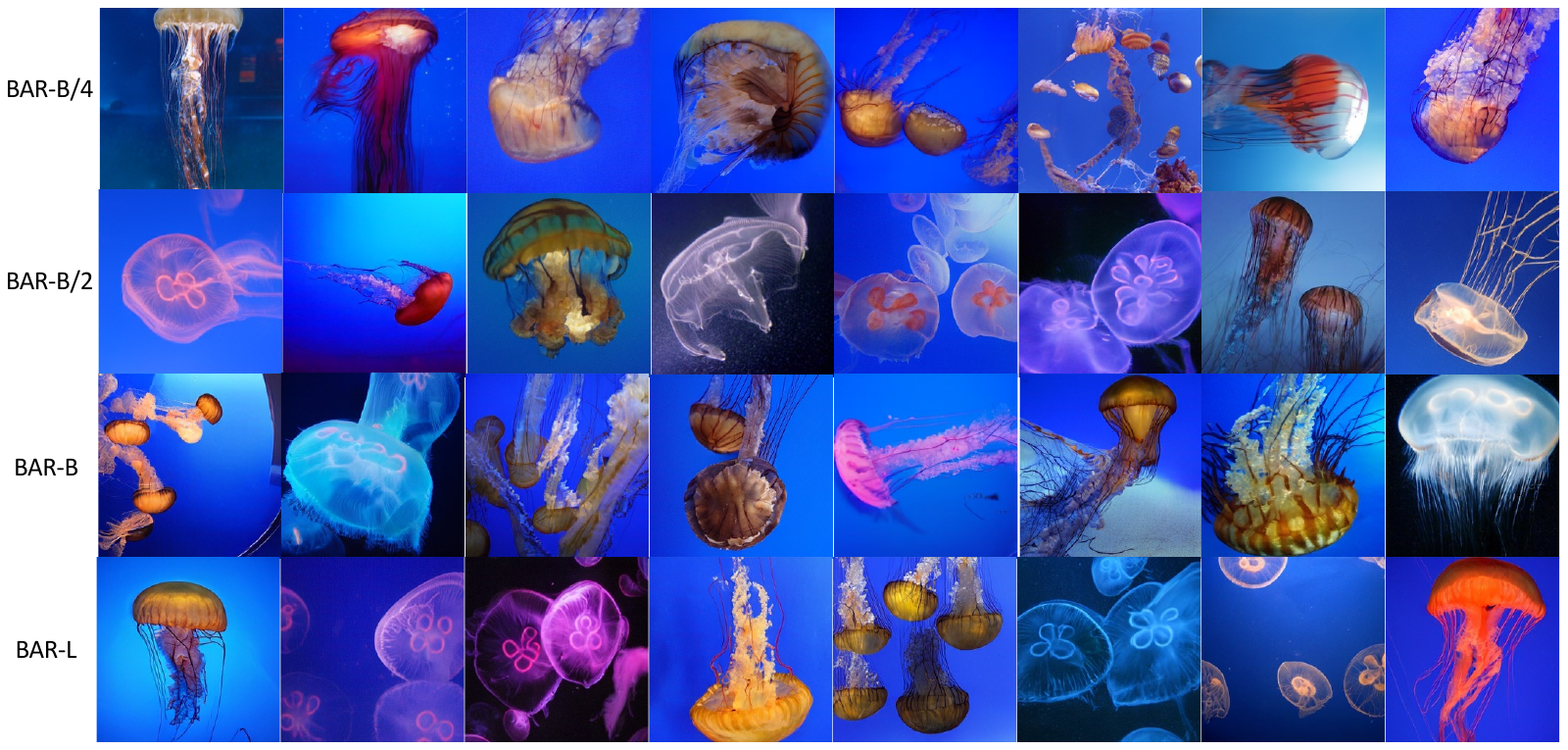}
    \vspace{-14ex}
    \caption{
    \textbf{Visualization samples from BAR models.} BAR is capable of generating high-fidelity image samples with great diversity. class idx 107: ``jellyfish''.
    }
    \vspace{-37ex}
    \label{fig:vis_more4}
\end{figure*}

\begin{figure*}[t]
    \centering
    \includegraphics[width=0.8\linewidth]{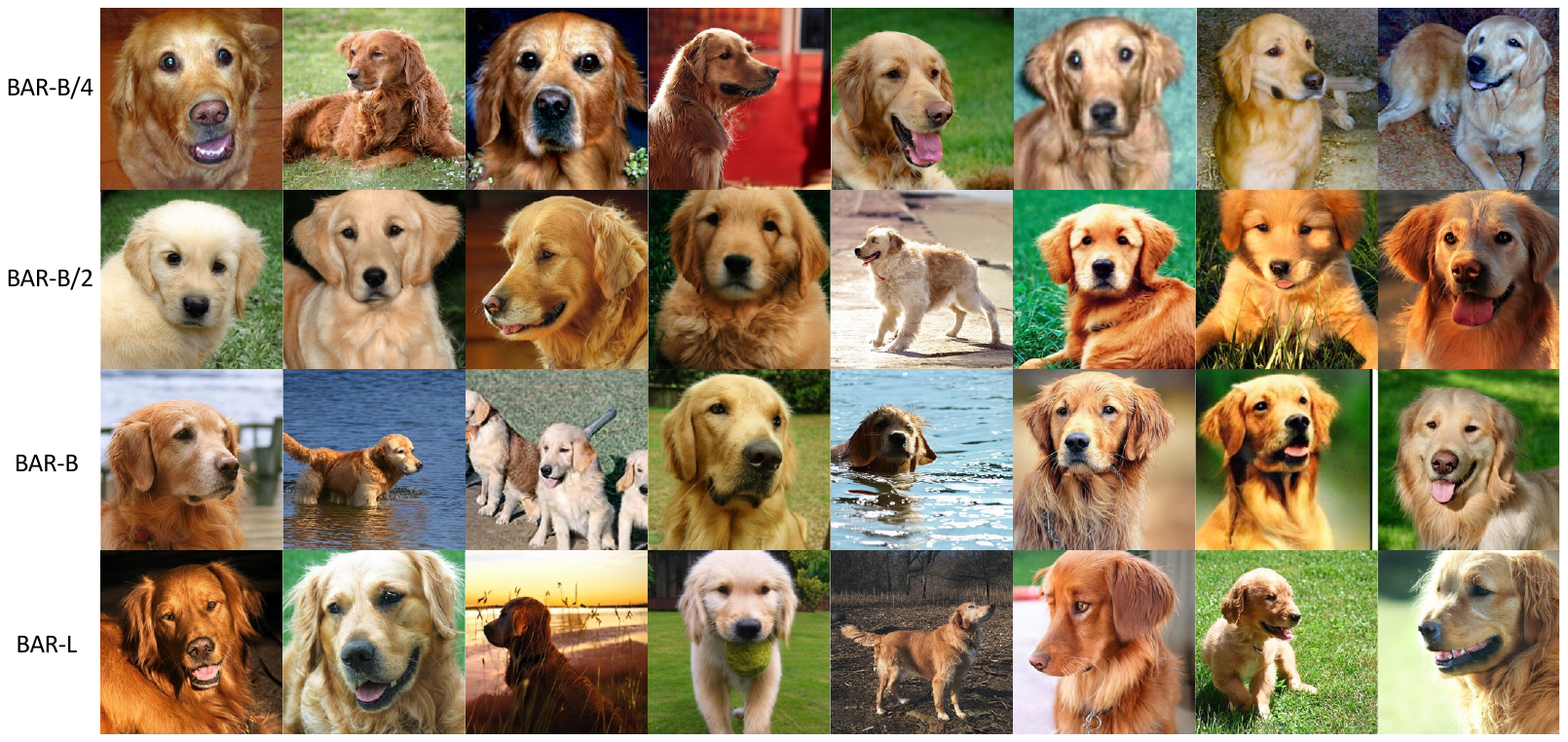}
    \vspace{-14ex}
    \caption{
    \textbf{Visualization samples from BAR models.} BAR is capable of generating high-fidelity image samples with great diversity. class idx 207: ``golden retriever''.
    }
    \vspace{-37ex}
    \label{fig:vis_more5}
\end{figure*}

\begin{figure*}[t]
    \centering
    \includegraphics[width=0.8\linewidth]{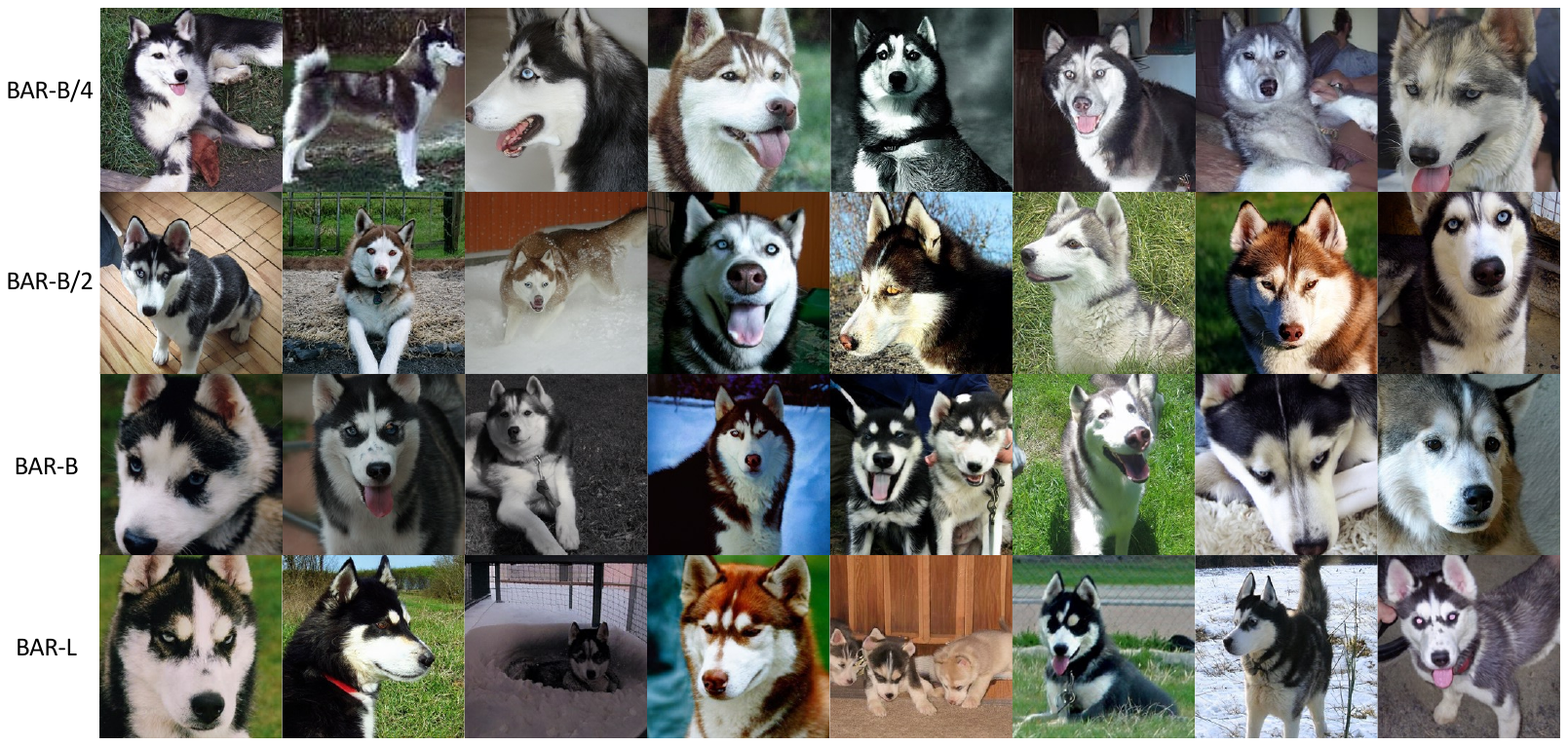}
    \vspace{-14ex}
    \caption{
    \textbf{Visualization samples from BAR models.} BAR is capable of generating high-fidelity image samples with great diversity. class idx 250: ``Siberian husky''.
    }
    \vspace{-15ex}
    \label{fig:vis_more6}
\end{figure*}

\begin{figure*}[t]
    \centering
    \vspace{-24ex}
    \includegraphics[width=0.8\linewidth]{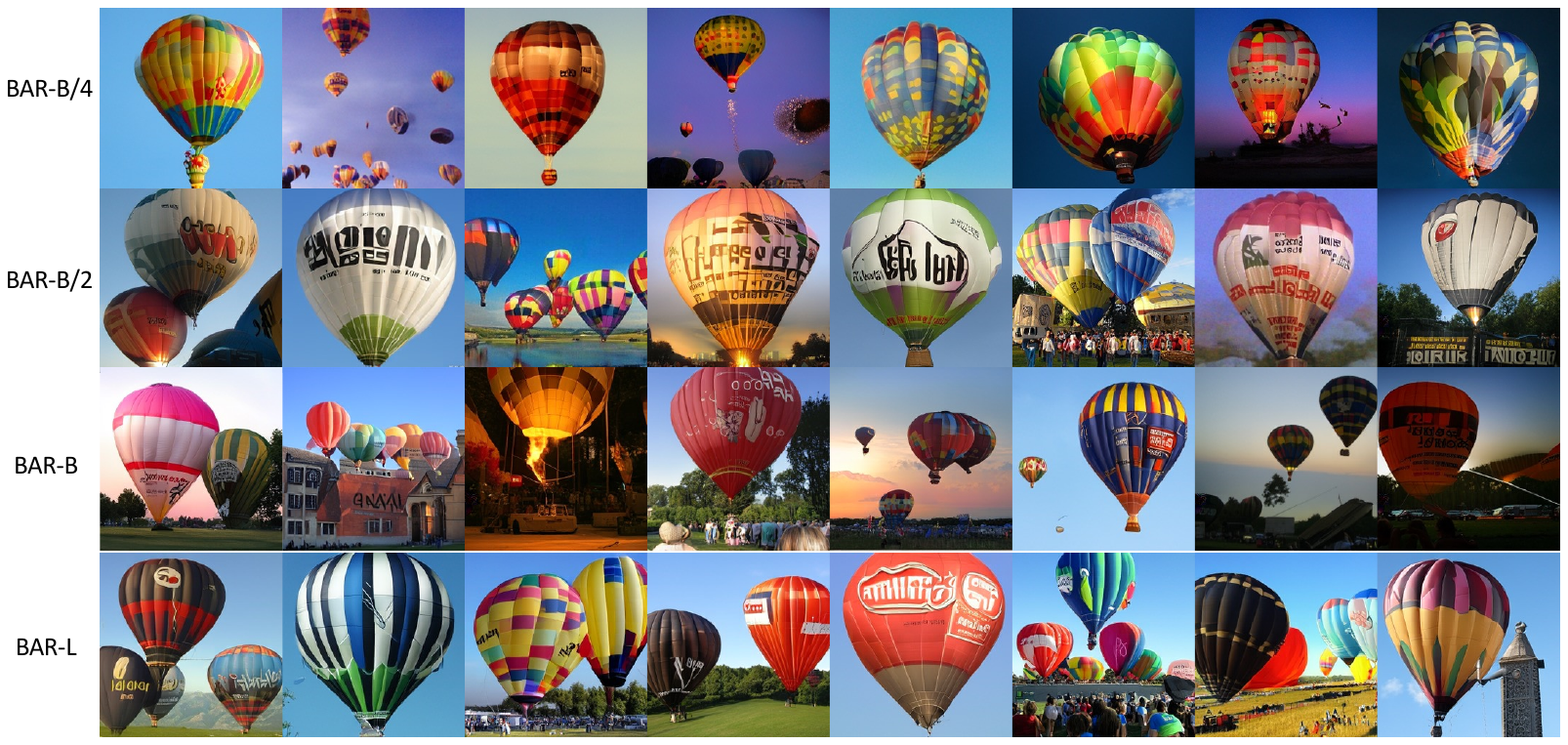}
    \vspace{-14ex}
    \caption{
    \textbf{Visualization samples from BAR models.} BAR is capable of generating high-fidelity image samples with great diversity. class idx 417: ``balloon''.
    }
    \vspace{-37ex}
    \label{fig:vis_more7}
\end{figure*}

\begin{figure*}[t]
    \centering
    \includegraphics[width=0.8\linewidth]{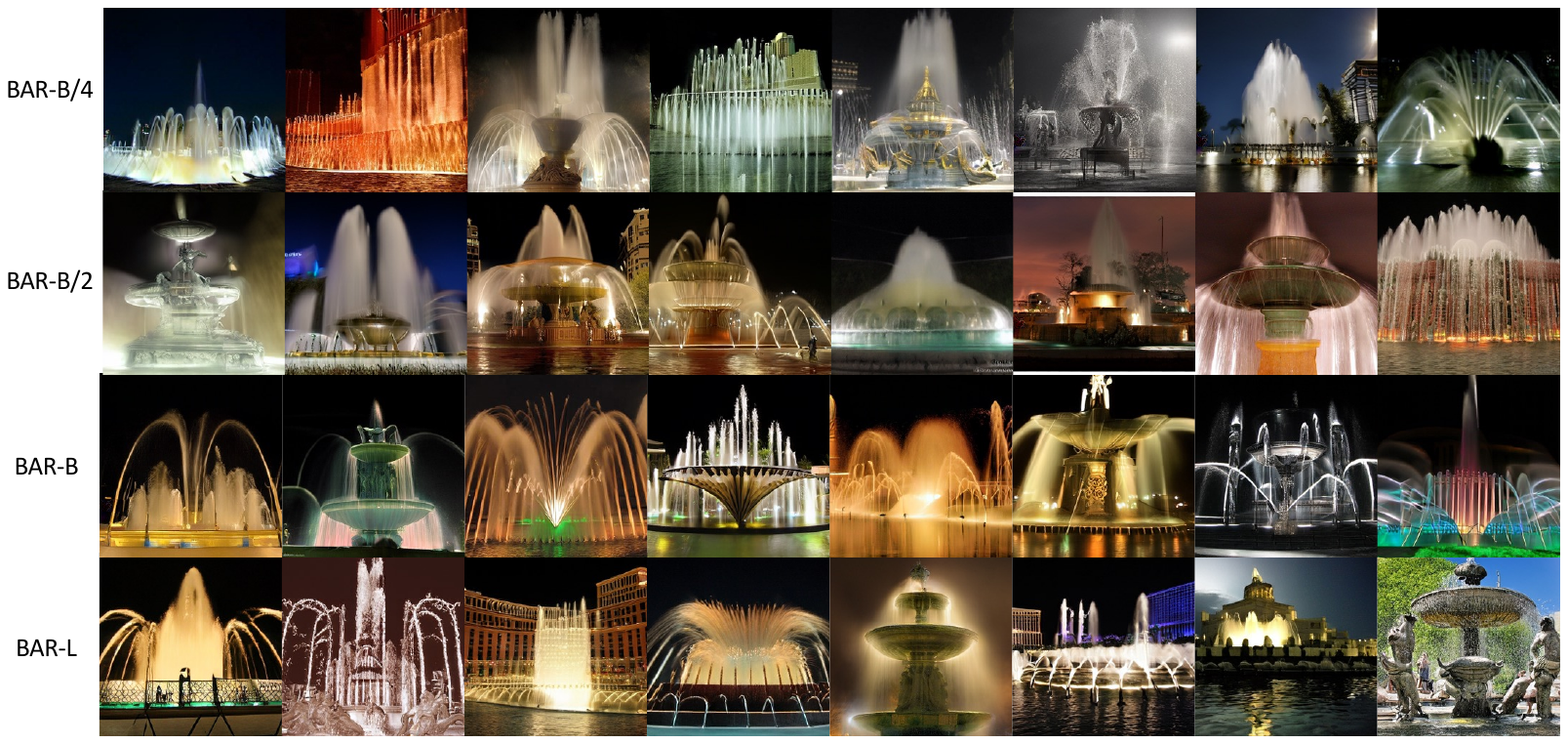}
    \vspace{-14ex}
    \caption{
    \textbf{Visualization samples from BAR models.} BAR is capable of generating high-fidelity image samples with great diversity. class idx 562: ``fountain''.
    }
    \vspace{-37ex}
    \label{fig:vis_more8}
\end{figure*}

\begin{figure*}[t]
    \centering
    \includegraphics[width=0.8\linewidth]{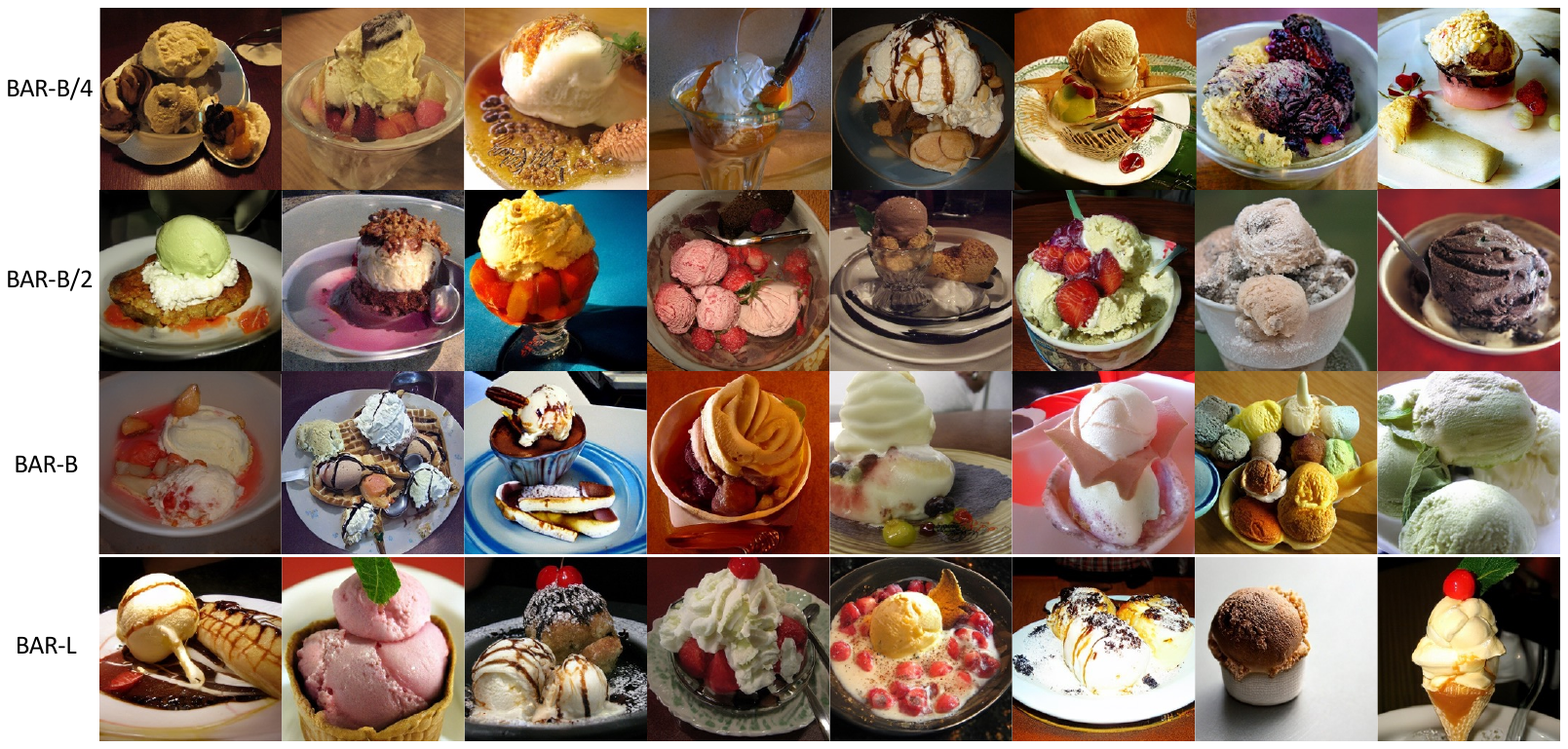}
    \vspace{-14ex}
    \caption{
    \textbf{Visualization samples from BAR models.} BAR is capable of generating high-fidelity image samples with great diversity. class idx 928: ``ice cream, icecream''.
    }
    \vspace{-15ex}
    \label{fig:vis_more9}
\end{figure*}

\begin{figure*}[t]
    \centering
    \vspace{-24ex}
    \includegraphics[width=0.8\linewidth]{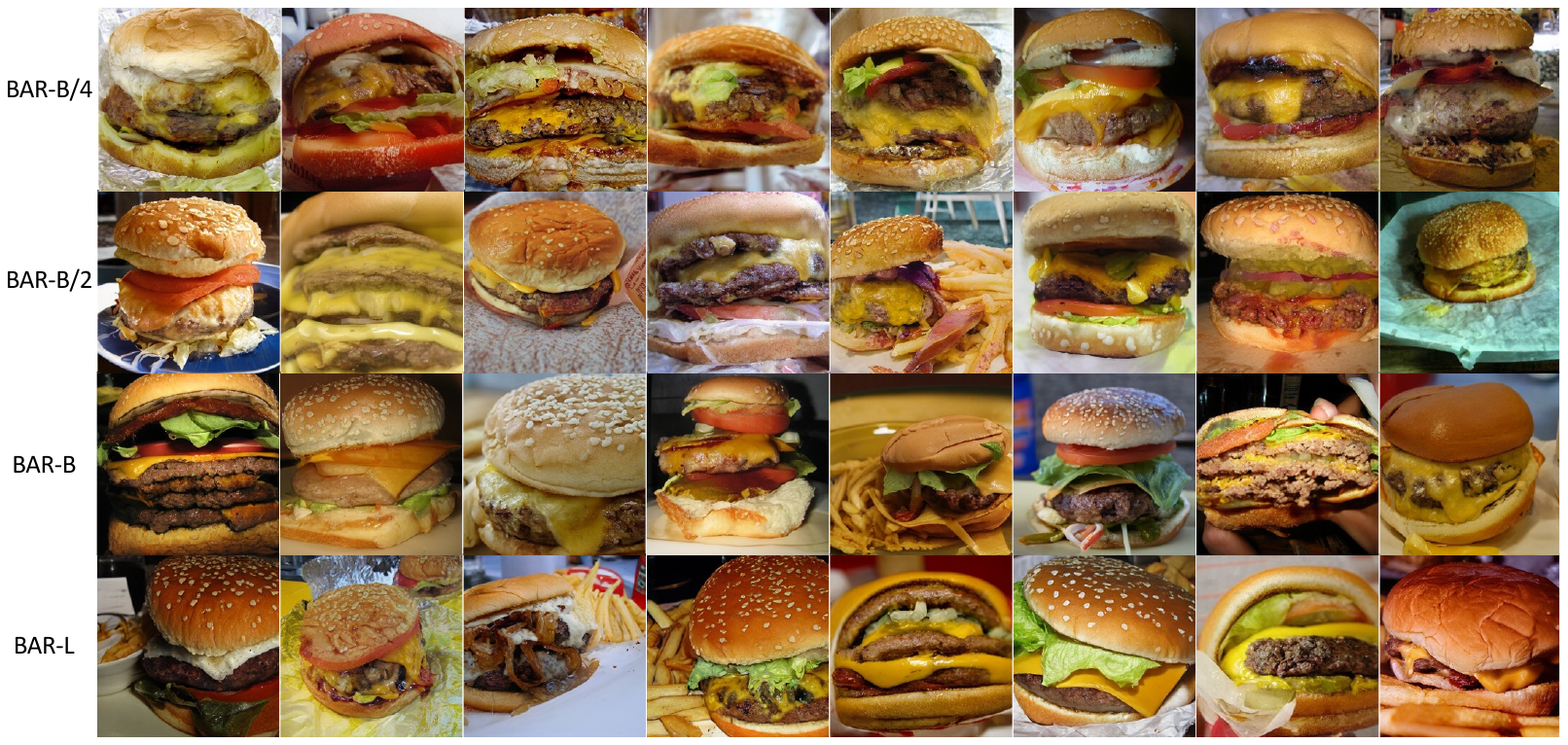}
    \vspace{-14ex}
    \caption{
    \textbf{Visualization samples from BAR models.} BAR is capable of generating high-fidelity image samples with great diversity. class idx 933: ``cheeseburger''.
    }
    \vspace{-37ex}
    \label{fig:vis_more10}
\end{figure*}

\begin{figure*}[t]
    \centering
    \includegraphics[width=0.8\linewidth]{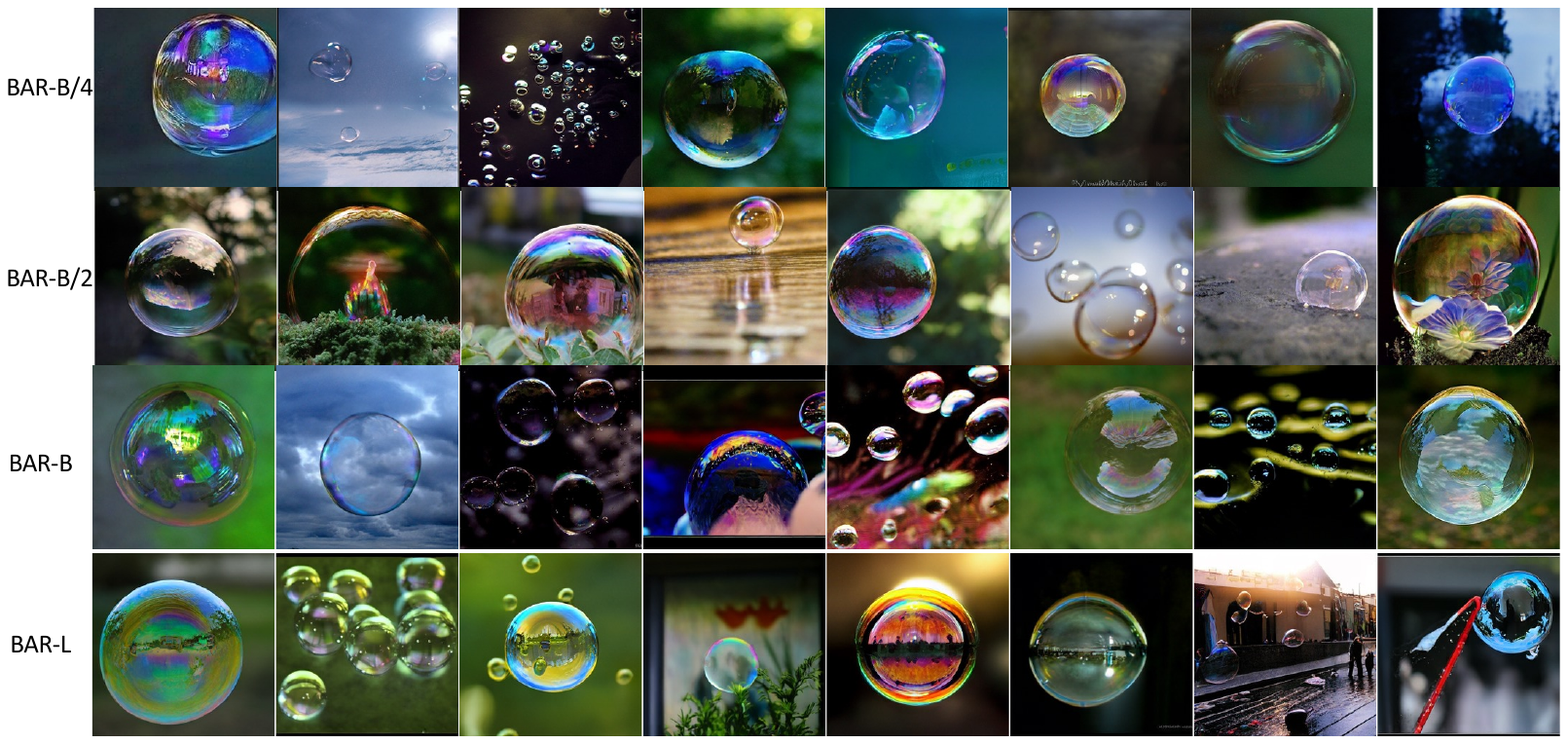}
    \vspace{-14ex}
    \caption{
    \textbf{Visualization samples from BAR models.} BAR is capable of generating high-fidelity image samples with great diversity. class idx 971: ``bubble''.
    }
    \vspace{-37ex}
    \label{fig:vis_more11}
\end{figure*}

\begin{figure*}[t]
    \centering
    \includegraphics[width=0.8\linewidth]{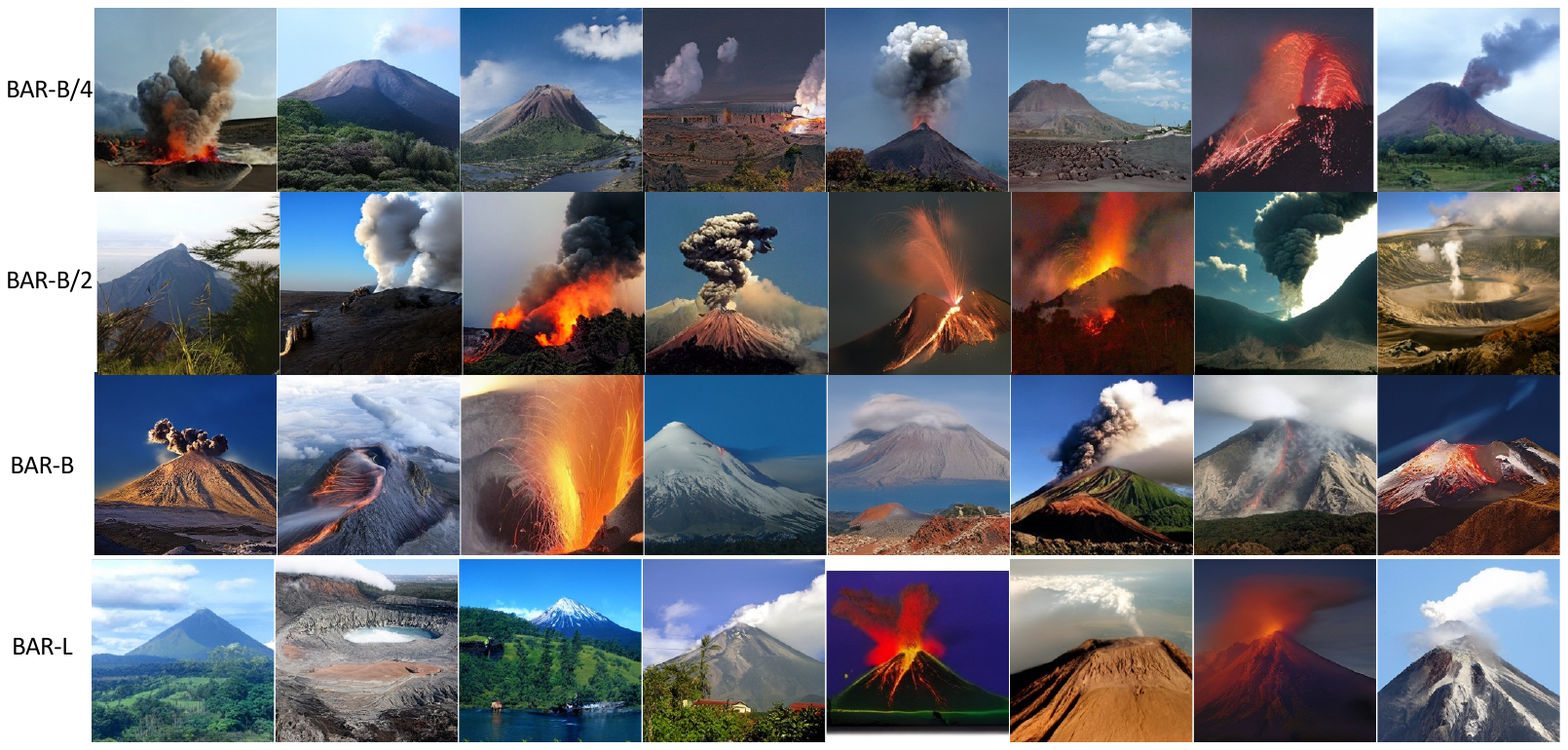}
    \vspace{-14ex}
    \caption{
    \textbf{Visualization samples from BAR models.} BAR is capable of generating high-fidelity image samples with great diversity. class idx 980: ``volcano''.
    }
    \vspace{-15ex}
    \label{fig:vis_more12}
\end{figure*}

We provide visualization results in \figref{fig:vis_more1}, \figref{fig:vis_more2}, \figref{fig:vis_more3}, \figref{fig:vis_more4}, \figref{fig:vis_more5}, \figref{fig:vis_more6}, \figref{fig:vis_more7}, \figref{fig:vis_more8}, \figref{fig:vis_more9}, \figref{fig:vis_more10}, \figref{fig:vis_more11}, and \figref{fig:vis_more12}.

\clearpage

\clearpage

\clearpage

\clearpage

\end{document}